\documentclass[12pt]{elsarticle}
\usepackage{graphicx}
\usepackage{amssymb}
\usepackage{multirow}
\usepackage{amssymb,amsmath}
\usepackage{makecell}
\usepackage{pgf,tikz,pgfplots}
\usepackage{url}
\usepackage{bm}
\usepackage{import}

\usepackage[top=2cm, bottom=2cm, left=2cm, right=2cm]{geometry}
\journal{  }

\usepackage{algorithm}
\usepackage{algorithmic}

\newtheorem{defn}{Definition}
\newtheorem{prop}{Proposition}
\newtheorem{example}{Example}

\begin{document}

\begin{frontmatter}

\title{RIGA: A Regret-Based Interactive Genetic Algorithm\footnote{This paper is an extension of work originally reported in~\cite{BenabbouAAAI20}. Compared to the original paper, we have considered another MOCO problem (the multi-objective knapsack) and another aggregation function (namely the Choquet integral). We have also added comparisons with a state-of-the-art evolutionary method (NEMO-II). }}
\author{Nawal Benabbou, Cassandre Leroy, Thibaut Lust}

\address{Sorbonne Université, CNRS, LIP6, F-75005 Paris, France \\ firstname.lastname@lip6.fr }

\begin{abstract}
In this paper, we propose an interactive genetic algorithm for solving multi-objective combinatorial optimization problems under preference imprecision. More precisely, we consider problems where the decision maker's preferences over solutions can be represented by a parameterized aggregation function (e.g., a weighted sum, an OWA operator, a Choquet integral), and we assume that the parameters are initially not known by the recommendation system. In order to quickly make a good recommendation, we combine elicitation and search in the following way:
1) we use regret-based elicitation techniques to reduce the parameter space in a efficient way, 2) genetic operators are applied on parameter instances (instead of solutions) to better explore the parameter space, and 3) we generate promising solutions (population) using existing solving methods designed for the problem with known preferences. 
Our algorithm, called RIGA, can be applied to any multi-objective combinatorial optimization problem provided that the aggregation function is linear in its parameters and that a (near-)optimal solution can be efficiently determined for the problem with known preferences. We also study its theoretical performances: RIGA can be implemented in such way that it runs in polynomial time while asking no more than a polynomial number of queries. The method is tested on the multi-objective knapsack and traveling salesman problems. For several performance indicators (computation times, gap to optimality and number of queries), RIGA obtains better results than state-of-the-art algorithms.
\end{abstract}

\begin{keyword}
Multiple objective programming \sep Interactive Method \sep Incremental Elicitation \sep Traveling Salesman Problem \sep Knapsack Problem.
\end{keyword}

\end{frontmatter}


\section{Introduction}

Many real world decision contexts require to take into account several conflicting and heterogeneous criteria evaluating quantitative issues (performance, duration, cost, etc.) and even sometimes qualitative issues (environmental aspects, customer opinions, etc.). Considering explicitly these criteria enables the decision-maker (DM) to appreciate the possible trade-offs and to progress towards the definition of a best
compromise solution. 

In this paper, we focus on the class of multi-objective combinatorial
optimization (MOCO) problems where the difficulty of considering multiple criteria is combined with the one resulting from the combinatorial structure of the set of feasible solutions. 
MOCO problems encompass a large variety of multi-objective variants of classical combinatorial optimization problems, which can be defined equivalently using a graph or an integer linear program, and whose single objective versions
($n=1$) are polynomial-time solvable (shortest path, minimum spanning tree, assignment, etc.) or
NP-hard (knapsack, set covering, traveling salesman, etc.). 
Formally, a MOCO problem can be defined as follows~\cite{Ehrgott05}: We are given $p$ finite discrete sets $X_i$ with $i \in P=\{1,\ldots, p\}$, from which is defined a set $\mathcal{X} \subseteq X_1 \times \ldots \times X_p$ of feasible solutions, and $n$ functions $y_j : \mathcal{X} \rightarrow \mathbb{R}$ with $j \in N=\{1,\ldots, n\}$, called objective functions or criteria (to be maximized or minimized). Any feasible solution $x \in \mathcal{X}$ is characterized by a performance vector $y(x)=(y_1(x), \ldots, y_n(x))$ such that $y_j(x)$ is the evaluation of $x$ with respect to 
criterion $j$ for any $j\in N$. In this setting, the following three problems are usually addressed: (1) finding the whole set of Pareto-optimal solutions\footnote{A solution is called Pareto-optimal if no other solution is better on all objectives while being strictly better on at least one objective.} (called the Pareto set hereafter) ; (2) determining a representative sample of the Pareto set; (3) identifying a good compromise solution according to the DM's preferences.

From a computational viewpoint, solving a MOCO problem raises several difficulties. First, the number of Pareto-optimal solutions often grows exponentially with the number of criteria in the worst case. Second, for most MOCO problems, the associated decision problem is NP-complete, even when the underlying single objective problem can be solved in polynomial time. Owing to these major computational difficulties, exact algorithms aiming to generate the Pareto set are often impractical, even for small-size problems. 
Moreover, in many real-world situations, the DM finds it more useful to be presented with a representative sample of candidate solutions rather than to be overwhelmed by a large list of solutions. 
In such situations, it seems more interesting to design algorithms constructing a ``well-represented'' Pareto set, i.e. a small subset of the Pareto set which ``covers'' the entire objective space. This approach is often implemented using algorithms based on a division of the objective space into different regions (e.g., \cite{KarasakalK09,Knowles00,Li2010,Lust14}) or on $\epsilon$-dominance (e.g., \cite{Laumanns2002}). However, in problems where the DM is able to give us some insights into the problem and is willing to share her preferences, it seems more appropriate to refine the Pareto dominance relation with preferences so as to discriminate between feasible tradeoffs and determine a single solution satisfying the subjective preferences of the DM. 

In the field of Multicritieria Decision Making, the DM's preferences are often represented by a \emph{parameterized} aggregation function which is used to compare alternatives by synthesizing their performances into overall utility values. 
Recently there has been a growing interest in rank-dependent aggregators such as Choquet integrals \cite{Grabisch10}, which enable to model complex decision behaviors by considering possible synergies between the objectives. They also make it possible to model preferences for ``well-balanced'' efficient solutions that cannot be obtained by optimizing a weighted sum. 
Although many algorithmic contributions focus on the elaboration of algorithms allowing the fast determination of the optimal solutions given an aggregation function, some others concern \emph{model-based} preference elicitation strategies aiming to quickly acquire the DM's preferences by specifying the \emph{preference parameters} modeling the relative importance of criteria. However, when the decision space is very large (which is the case in MOCO problems), preference elicitation is often a challenging issue since usual preference elicitation methods based on
systematic pairwise comparisons are often ineffective. 
Here we address both issues simultaneously by studying the integration of preference elicitation into the resolution of MOCO problems with rank-dependent aggregation functions.

Preference elicitation on combinatorial domains is an active topic that has motivated several contributions in various contexts, e.g., in multi-objective state space search \cite{benabbouPernyAaai2015}, in stable matching problems \cite{DrummondB14}, in constraint satisfaction problems \cite{boutilier-minimaxAIJ06}, in Markov Decision Processes \cite{ReganB11,WengP13} and in MOCO problems \cite{benabbouPerny2018}.
Among preference elicitation methods, incremental approaches are of special interest because they aim to analyze the set of feasible solutions in order to identify the critical preference information needed to find the optimal alternative. By a careful selection of preference queries, they make it possible to determine the optimal choice within large sets, using a reasonably small number of questions (see e.g., \cite{Chajewska00makingrational} for an example in a bayesian setting). 
Here we focus on \emph{regret-based} incremental preference elicitation as it was proved to be very effective in reducing the imprecision attached to preference parameters in practice (see e.g., \cite{boutilier-minimaxAIJ06,WangBoutilier03}). 
The idea is to use the minimax regret decision criterion to select informative preference queries at each step of the elicitation process, so as to progressively reduce the set of admissible parameters until a \emph{robust} recommendation can be made. This elicitation approach has been mainly studied with linear aggregators in non-combinatorial problems.

For MOCO problems with linear aggregators, it has been recently proposed to combine search and regret-based incremental elicitation by asking preference queries during the construction of the (near-)optimal solution \cite{benabbouPernyAaai2015}. The basic principle consists in constructing the optimal solution from optimal sub-solutions using the available preference information, asking new preference information only when necessary.  The motivation behind is to save computation time by using the DM’s answers to quickly focus the search on the most promising solutions and to reduce the elicitation burden by generating preference queries only to discriminate between (sub-)solutions found during the search. 
In this paper, we show how to efficiently extend this approach to MOCO problems with complex aggregation functions. The main novelty is to make use of metaheuristics instead of constructive algorithms, which enables to tackle preference-based combinatorial optimization problems for which no efficient exact algorithm is known. More precisely, we introduce an interactive genetic algorithm that uses incremental elicitation techniques to select individuals from populations. The combination of preference elicitation and genetic algorithms has several specific advantages. First, preference queries only involve complete feasible solutions. This provides at least two advantages: 1) solutions are easier to compare, and 2) no independence assumption is required in the definition of preferences. The latter point is of special interest when preferences are represented by {\em non-linear} aggregators because the performance of sub-solutions is often a poor predictor of the actual performance of their extensions. 
Another interest of combining regret-based incremental elicitation and genetic algorithms is to produce interactive methods with performance guarantees: the proposed algorithm
1) generates no more than a polynomial number of queries when the population size and number of generations are  polynomial (as preference information is only used to discriminate between solutions within a population), and 2) runs in polynomial time when considering aggregation functions such that the minimax regret can be computed efficiently, and the MOCO problem can be solved (exactly or approximately) in polynomial time when the aggregation function is known.

The proposed interactive genetic algorithm, called RIGA for Regret-based Interactive Genetic Algorithm, follows the traditional scheme of genetic algorithms but differs in the following way: Genetic operators are applied on parameter instances (not on solutions) which allows to explore the parameter space in an efficient way. For every generated parameter instance, a feasible solution is obtained by applying an existing efficient solving method designed for the considered MOCO problem with known preferences. The selection step is then performed by comparing the generated solutions using standard regret-based incremental elicitation techniques, which allows to efficiently reduce the parameter space during the resolution. 
The proposed interactive method is general in the sense that it can be applied to any MOCO problem provided that it can be efficiently solved (exactly or approximately) when the aggregation function is known, and that minimax regrets can be efficiently computed with the aggregation function under consideration.

The paper is organized as follows: Firstly, we conduct a literature overview on interactive genetic algorithms and regret-based solving methods. Secondly, we recall some useful notions related to rank-dependent aggregation functions and the general principle of incremental elicitation driven by the minimax regret decision criterion. Thirdly, a new regret-based interactive genetic algorithm is proposed and illustrated on a small instance of the multi-objective knapsack problem. Finally, we provide numerical tests showing its practical efficiency on two multi-objective problems, namely the knapsack and traveling salesman problems, comparing its performances with that of several existing methods.

\section{Related Work}

The new method is based on two distinct concepts: interactive evolutionary methods and regret-based solving methods. Accordingly we divided the related works into two parts. 

\subsection{Interactive evolutionary algorithm}

During these past few years, integrating preferences into evolutionary algorithms has become increasingly popular, see e.g. \cite{Branke2008,Xin2018} for some surveys.
Due to the high number of different interactive methods that has been developed, Xin \emph{et al}~\cite{Xin2018} have recently established a taxonomy identifying the important factors to differentiate these methods. Four essential design factors are defined: interaction pattern (how the interaction with the DM is scheduled during the run), preference information (how the preference information is obtained from the DM), preference model (utility function, dominance relation or decision rules), and search engine (how the interesting solutions are produced, e.g. mathematical programming techniques or heuristics).

For the new method developed in this paper, the interaction with the DM is done during the run, the preference information is retrieved from pairwise comparisons, the preference model is a utility/aggregation function and the search engine can be both mathematical programming techniques or heuristics. 
To the best of our knowledge, the existing methods that share the same factors are: the Interactive Evolutionary Metaheuristic (IEM)~\cite{Phelps2003}, the Interactive Pareto Memetic Algorithm (IPMA)~\cite{Jaszkiewicz2004}, the Progressively Interactive Evolutionary Multi-Objective approach using Value Functions (PI-EMOVF)~\cite{Deb2010}, the Necessary-preference-enhanced Evolutionary Multi-objective Optimizer (NEMO)~\cite{Branke2015}, the Brain-Computer Evolutionary Multi-objective Optimization Algorithm (BC-EMOA)~\cite{BattitiP10} and the Interactive Non-dominated Sorting algorithm with Preference Model called INSPM~\cite{Pedro2014}.  
All but the first two of these methods are adaptations of NSGA-II~\cite{Deb2002}, which is the reference algorithm for solving continuous multi-objective problems, and they have not been tested in MOCO problems. 
The first two methods only consider utility functions based on linear or Chebyshev aggregation. 
In IEM, linear programming techniques are used to learn the parameters of the utility function, whereas an internal genetic algorithm is employed in IPMA. In~\cite{Jaszkiewicz2004}, IEM and IPMA were directly compared and IPMA achieved better results on the considered multi-objective traveling salesman problem instances. 
IPMA follows the classical steps of genetic algorithms while generating pairwise comparison queries periodically to reduce the set of possible utility functions. 
The frequency of preference queries is controlled by a comparison probability, which is progressively reduced during the search. Instead, our algorithm uses regret-based incremental elicitation techniques to select informative preference queries and generates promising solutions during the search. As a result,
our method generates at most 25 queries on existing traveling salesman instances with 300 cities and 7 objectives (see the numerical section) whereas IPMA needs between 30 and 60 queries on smaller instances (150 cities and 6 objectives). Moreover, we propose to apply the crossover and mutation operators on the preference parameters instead of solutions to better cover the space of admissible utility functions.


In the numerical section, we will compare the results obtained by our algorithm to that of a new version of NEMO called NEMO-II~\cite{BrankeCGSZ16}. NEMO follows the same scheme as NSGA-II, except that the dominance relation used in the sorting step is replaced by a necessary preference relation which is built from the available preference information (expressed in terms of pairwise comparisons of solutions): solution $a$ is necessarily preferred to solution $b$ if and only if $a$ is at least as good as $b$ for all value functions compatible with the available preference data. The main difference between NEMO and NEMO-II lies in the computation of the necessary preference relation: NEMO requires to solve a quadratic number of linear programs in the worst-case whereas NEMO-II only performs a linear number of such optimizations. 
Moreover, NEMO-II is able to handle inconsistencies in the information provided by the DM: if at some point there is no value function compatible with the collected preference information, then the constraints related to the oldest pairwise comparisons are removed until feasibility is restored. NEMO-II is also able to switch from a simple preference model (a weighted sum) to a more sophisticated one (the 2-additive Choquet integral) in order to capture more complex decision behaviors.

\subsection{Regret-based solving methods}

The first regret-based incremental elicitation procedures designed for MOCO problems are constructive methods interweaving search and preference elicitation by asking preference queries during the construction of the optimal solution \cite{benabbouPernyAaai2015}. 
In this paper, we explore another way by considering metaheuristics instead of constructive algorithms so as to be able to tackle preference-based combinatorial optimization problems for which no efficient exact algorithm is known (e.g., the multi-objective traveling salesman problem with preferences represented by a Choquet integral). 

More recently, it has been proposed to combine search and regret based incremental elicitation in a different way \cite{BenabbouL19SUM,BenabbouL19}. The idea is to identify informative queries by exploiting the extreme points of the polyhedron representing the admissible preference parameters. This method, called IEEP for Incremental Elicitation based on Extreme Points, proceeds as follows: At each iteration step, promising solutions are generated using the extreme points of the polyhedron, the DM is asked to compare two of these solutions, and the polyhedron is updated according to the collected preference information. These extreme points are also used to provide a stopping criterion guaranteeing that the returned solution is optimal (or near-optimal) according to the DM's preferences. Compared to the genetic algorithm proposed in this paper, IEEP is an exact exponential time method that may generate an exponential number of queries, whereas the proposed method is a poly-time heuristic that asks no more than a polynomial number of queries. These two procedures will be compared in the numerical section.


Note that regret-based solving methods have also been extensively used in robust combinatorial optimization. In this setting, only one objective function is considered, but it may be uncertain or imprecise. The concept of scenario is then used to model the uncertainty/imprecision over the objective function. The goal is to find a solution that is optimal according to the min-max regret criterion, i.e. a solution minimizing the maximum deviation, over all possible scenarios, between the value of the solution and the optimal value of the corresponding scenario. This approach has been studied for several combinatorial optimization problems, e.g. the shortest path, the spanning tree, the assignment, the knapsack, the $s-t$ cut, see~\cite{Aissi2009} for a survey.

\section{Background and Notations}

\subsection{Multi-Objective Combinatorial Optimization}

In this paper, we consider MOCO problems with $n$ objective functions $y_j,j \in N= \{1,\ldots,n\}$, to be minimized (e.g., the traveling salesman, knapsack, minimum spanning tree problems). These optimization problems can be formulated as follows: 
$$ `` \underset {x \in \mathcal{X}} {\textrm{minimize}} \textrm{''} \big(y_1(x),\ldots,y_n(x)\big)$$
where $\mathcal{X}$ is the \emph{feasible set} in the decision space, typically defined by some constraint functions characterizing the feasible solutions. For example, for the traveling salesman problem, $\mathcal{X}$ is the set of all Hamiltonian cycles of the graph, whereas for the knapsack problem, $\mathcal{X}$ only includes the subsets of items satisfying the knapsack capacity constraint. 
In MOCO problems, any feasible solution $x \in \mathcal{X}$ is associated with a cost/performance vector $y(x)= (y_1(x),\ldots,y_n(x)) \in  \mathbb{R}^n$ where $y_j(x)$ is the evaluation of $x$ on the $j$-th criterion/objective. In MOCO problems, solutions are usually compared through their images in the objective space (also called \emph{points}) using the \emph{Pareto dominance} relation: we say that point $u=(u_1,\ldots,u_n) \in \mathbb{R}^n$ Pareto dominates point $v=(v_1,\ldots,v_n)\in \mathbb{R}^n$  (denoted by $u \prec_P v$) if and only if $u_j \leq v_j$  for all $j \in \{1,\ldots,n\}$, with at least one inequality being strict. Solution $x^* \in \mathcal{X}$ is called \emph{efficient} (or \emph{Pareto-optimal}) if there does not exist any other feasible solution $x \in \mathcal{X}$ such that $y(x) \prec_P y(x^*)$ (its image in the objective space is called a \emph{non-dominated} point). 


In this paper, we assume that the DM's preferences over solutions can be represented by a parameterized scalarizing function $f_\omega$ such that solution $x\in \mathcal{X}$ is preferred to solution $x'\in \mathcal{X}$ if and only if $f_\omega(y(x)) < f_\omega(y(x'))$. The goal is therefore to identify a feasible solution $x$ that minimizes the aggregated value $f_\omega(y(x))$. 
We assume here that $f_\omega$ is linear in its parameters $\omega$. For example, function $f_\omega$ can be a weighted sum, i.e. $f_\omega(y(x)) = \sum_{j=1}^n \omega_j y_j(x)$ where $\omega=(\omega_1,\ldots, \omega_n)$ is a normalized weighting vector such that $\omega_j \in [0,1]$ is the weight attached to objective $j \in N$. Recently there has been a growing interest in rank-dependent aggregators, which enable to model more complex decision behaviors. Rank-dependent aggregation functions are scalarizing functions that sort the performance values by increasing order before mapping the performance vector into a scalar value, thus enabling to assign weighting coefficients to ranks rather than to objectives. From a descriptive and prescriptive viewpoint, the interest of resorting to such aggregation functions is twofold: 1) It provides a more general and flexible class of decision models that can be tuned to the observed preferences, and 2) it enables to enhance the possibility of finding good compromise solutions within the Pareto set. The latter point is illustrated in the following optimization example. 

\begin{example}
Let us consider a bi-objective optimization problem where the set of feasible solutions is represented in the objective space in Figure \ref{figEx1}. 
In this figure, every point represents a feasible cost vector and the red points represent the Pareto set. As can be seen from the convex hull of these points, only two points in the Pareto set can be obtained by minimizing a weighted sum of the costs, namely $A$ and $B$. However, more interesting compromise solutions can be found when considering rank-dependent aggregators such as Choquet integrals \cite{Grabisch10}, as can be seen from the isopreference curve represented by the dashed lines.


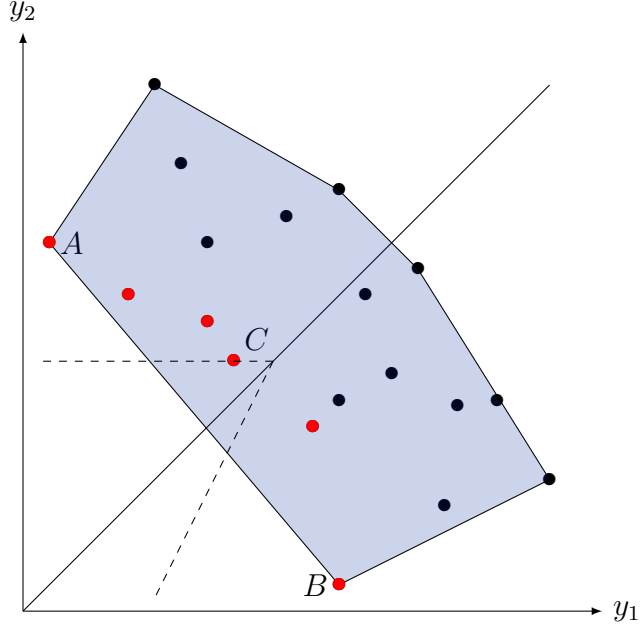
\begin{figure}[h]
    \centering
\begin{tikzpicture}[scale=0.35]
 
\coordinate (A) at (0,0);
\coordinate (B) at (0,22);
\coordinate (D) at (22,0);

\draw[-latex] (A) -- (B);
\draw[-latex] (A) -- (D);
\draw (A) -- (20,20);

\draw (D) node[right]{$y_1$};
\draw (B) node[above]{$y_2$};

\coordinate (X) at (20,5);
\coordinate (Y) at (5,20);
\coordinate (Z) at (12,11);

\coordinate (W) at (8,9.5);
\coordinate (V) at (11,7);

\coordinate (E) at (1,14);
\coordinate (I) at (12,1);

\coordinate (G) at (12,16);
\coordinate (J) at (15,13);

\coordinate (H) at (4,12);
\coordinate (F) at (16,4);

\coordinate (L) at (7,11);
\coordinate (Q) at (12,8);
\coordinate (O) at (7,14);

\coordinate (P) at (10,15);
\coordinate (K) at (13,12);
\coordinate (M) at (14,9);
\coordinate (N) at (18,8);
\coordinate (R) at (16.5,7.8);
\coordinate (U) at (6,17);

\draw (E) node[right]{$A$};
\draw (W) node[above right]{$C$};
\draw (I) node[left]{$B$};
\draw (W) node{$\bullet$};
\draw (V) node{$\bullet$};
\draw (E) node{$\bullet$};
\draw (F) node{$\bullet$};
\draw (G) node{$\bullet$};
\draw (H) node{$\bullet$};
\draw (I) node{$\bullet$};
\draw (J) node{$\bullet$};
\draw (K) node{$\bullet$};
\draw (L) node{$\bullet$};
\draw (M) node{$\bullet$};
\draw (N) node{$\bullet$};
\draw (O) node{$\bullet$};
\draw (P) node{$\bullet$};
\draw (Q) node{$\bullet$};
\draw (R) node{$\bullet$};
\draw (X) node{$\bullet$};
\draw (Y) node{$\bullet$};
\draw (U) node{$\bullet$};

\fill[color=green!20!blue!80!black, opacity=0.2] (E) -- (Y) -- (G) -- (J) -- (X) -- (I)--cycle;
\draw (E) -- (Y) -- (G) -- (J) -- (X) -- (I) --cycle;

\draw (W) node[color=red]{$\bullet$};
\draw (E) node[color=red]{$\bullet$};
\draw (I) node[color=red]{$\bullet$};
\draw (H) node[color=red]{$\bullet$};
\draw (L) node[color=red]{$\bullet$};
\draw (V) node[color=red]{$\bullet$};

\draw[dashed] (9.5,9.5) -- (0.5,9.5);
\draw[dashed] (9.5,9.5) -- (5,0.5);

\end{tikzpicture}
    \caption{Optimum obtained with a Choquet integral \label{figEx1}}
\end{figure}

\end{example}

In this paper, we focus on two families of rank-dependent aggregators: ordered weighted averages and Choquet integrals. 
To specify such aggregation functions, we will often denote by $(\cdot)$ a permutation defined on $\{1,\ldots,n\}$ such that $y_{(1)}(x) \leq ... \leq y_{(n)}(x)$, sorting the performance values of a given solution $x$ from the smallest to the largest.

\subsection{Ordered Weighted Averages}

Ordered Weighted Averages (OWA) is one of the simplest families of rank-dependent aggregation functions: it is simply defined as the weighted sum applied on the sorted performance vectors  \cite{Yager88}. OWA is specified by a normalized weighting vector $\omega = (\omega_1, \ldots, \omega_n) \in [0,1]^n$ where $\omega_j$ is the weight associated to the performance ranked in the $j$-th position. Formally, for any solution $x \in \mathcal{X}$, the OWA value is defined by:

\begin{defn}[OWA Operator]
The OWA value of solution $x \in \mathcal{X}$ is:
$$ \mbox{OWA}(x,\omega) = 
f_\omega(y(x)) = \sum_{j=1}^n \omega_j y_{(j)}(x)
$$
\end{defn}

 This family of aggregators includes the minimum, maximum, median and all order statistics as a special case. Note that $\mbox{OWA}(x,\omega)$ is not linear in $x$ for fixed $\omega$ but it is linear in $\omega$ for fixed $x$ (this property will be useful when computing max regrets later). Note also that OWA operators are symmetric as the components of the performance vectors are sorted before computing the aggregated value. In particular, this property seems natural when the objectives are individual points of view in a collective decision problem. OWA is often used in social choice theory with decreasing weights in order to favor well-balanced Pareto-optimal solutions (see e.g., \cite{KorhonenP90,LescaP10,Ogryc00} in fair allocation problems or \cite{goldsmith2014voting,elkind2015owa} in voting problems). In this paper, we consider optimization problems where an OWA operator must be minimized (as it is applied to cost vectors instead of utility vectors). In minimization problems, we need to use increasing weights to favor well-balanced solutions, due to the following property:

\begin{prop} \label{pigouowa} Let $u = (u_1,\ldots,u_n) \in \mathbb{R}^n$ be a real-valued vector such that there exist $j, k \in N$ satisfying $u_k < u_j$ with $k<j$. For any $\varepsilon \in (0, u_j - u_k)$, let $u^\varepsilon = (u_1, \ldots, u_j -\varepsilon, \ldots, u_k + \varepsilon, \ldots, u_n)$ be the real-valued vector obtained from $u$ by performing a transfer of size $\varepsilon$ from component $j$ to component $k$. We have:
$$
\Big ( \forall \ell \in \{1,\ldots, n-1\}, \omega_\ell \le \omega_{\ell+1} \Big ) \Rightarrow \mbox{OWA}(u, \omega) \ge \mbox{OWA}(u^\varepsilon, \omega)
$$
\end{prop}
In other words, when using increasing weights, reducing the inequality between two objectives always decreases the OWA value; this is due to the fact that we give more importance to the worst performance, less importance to the second worst performance, and so on. 
Such transfers are known as Pigou-Dalton transfers in Economics (see e.g., \cite{Weymark81}). Besides this property, OWA operators are monotonically increasing with respect to every component. From these two properties, we can conclude that the solutions minimizing an OWA operator are those Pareto-optimal solutions that cannot be improved in terms of Pigou-Dalton transfer. Thus minimizing an OWA operator with increasing weights helps promoting balanced solutions while ensuring overall efficiency in minimization problems.

\subsection{Choquet Integrals}

Choquet Integrals \cite{Grabisch10,choquet53,Sch86} is a more general family of aggregators which is really appealing for preference modeling because it enables to model different kinds of interactions between the objectives. More precisely, they perform a weighted aggregation of objective values using a (normalized) capacity function $\omega:2^N \rightarrow [0,1]$ which assigns weights to coalitions of objectives, allowing to model positive and/or negative interactions among objectives, covering an important range of possible decision behaviors.  Formally, a capacity function is defined by: 
\begin{defn}[Capacity Function]
Function $\omega:2^N \rightarrow [0,1]$ is a (normalized) capacity function if and only if:
\begin{itemize}
    \item $\omega(\emptyset)=0$, $\omega(N) = 1$ (normalization),
    \item and $\omega(A) \le \omega(B)$ for all $A \subset B \subseteq N$ (monotonicity).
\end{itemize}
\end{defn}

\noindent A capacity $\omega$ is said to be:
\begin{itemize}
    \item convex or supermodular when $\omega(A\cup B)+\omega(A\cap B)\geq \omega(A)+\omega(B)$ for all $A, B\subseteq N$,
    \item additive when $\omega(A\cup B)+\omega(A\cap B) = \omega(A)+\omega(B)$ for all $A, B\subseteq N$,
    \item and  concave or submodular when $\omega(A\cup B)+\omega(A\cap B)\leq \omega(A)+\omega(B)$ for all $A, B\subseteq N$.
\end{itemize}
We now use the notion of capacity to define the Choquet integral model. 
\begin{defn}[Choquet integral] The Choquet value of solution $x \in \mathcal{X}$ is: $$Choquet(x,\omega) = f_\omega(y(x)) = \sum_{j=1}^n \Big (y_{(j)}(x) - y_{(j-1)}(x)\Big ) \omega(X_{(j)}) \mbox{ with } y_{(0)}(x) = 0$$
where $X_{(j)} =  \{(j),\ldots, (n)\}$ is the set of objectives with respect to which $x$ has a cost greater or equal to $y_{(j)}(x)$.
\end{defn}
 Note that $Choquet(x,\omega)$ is not linear in $x$ for fixed $\omega$ but it is linear in $\omega$ for fixed $x$. 
 For illustration purposes, let us consider the following simple example. 
 
 \begin{example}
 Consider the following capacity function:
 
 \begin{center}
     \noindent $\begin{array}{ccccccccc}
\hline
  & \emptyset & \{1 \} & \{2 \} & \{3 \} & \{1,2 \} & \{1,3 \} & \{2,3 \} & \{1, 2, 3 \} \\
\hline
\omega & 0 & 0.2 & 0.1 & 0.3 & 0.4 & 0.7 & 0.6 & 1 \\
\hline
\end{array}$\\[1ex]
 \end{center}
 For a solution $x$ with cost vector $y(x)=(3,2,5)$, the Choquet integral value is $f_\omega(y(x)) = (2-0)\times \omega(\{1,2,3\}) + (3-2)\times \omega(\{1,3\})+ (5-3) \times \omega(\{3\}) = 2\times 1 + 1\times 0.7+ 2 \times 0.3 = 3.3$. 
 For  a solution $x'$ whose cost vector is $y(x')=(1,4,3)$, the Choquet integral value is $f_\omega(y(x')) = (1-0)\times \omega(\{1,2,3\}) + (3-1)\times \omega(\{2,3\})+ (4-3) \times \omega(\{2\}) = 1\times 1 + 2\times 0.6+ 1 \times 0.1 = 2.3$. Here we have $f_\omega(y(x')) < f_\omega(y(x))$, which means that the DM strictly prefers solution $x'$ to solution $x$.
 \end{example}
 
 In the definition of the Choquet integral model, the use of a capacity $\omega$ instead of an arbitrary set-function enforces compatibility with Pareto-dominance due to the monotonicity of $\omega$ with respect to set inclusion. In other words, it ensures that inequality $Choquet(x,\omega) \le Choquet(x',\omega)$ is satisfied whenever $y(x) \prec_P y(x')$ holds.

The family of Choquet integrals includes many aggregators as special cases. For example, it boils down to the family of weighted sums when considering additive capacities and it corresponds to the family of OWA aggregators when using symmetric capacities (i.e., capacities such that there exists some non-decreasing function $\psi$ satisfying $\omega(A) = \psi(|A|)$ for all $A \subseteq N$). 

Choquet integrals are known to be convex whenever $\omega$ is concave (submodular) and concave whenever $\omega$ is convex (supermodular) \cite{Lovasz83}. In minimization (resp. maximization) problems, the use of a concave (resp. convex) capacity allows to model preferences for well-balanced solutions, as shown in the following proposition \cite{ChaTa99}:

\begin{prop} Let $\omega$ be a concave capacity. 
For all $x^1, \ldots, x^p \in \mathbb{R}^n$ and for all $\lambda_1,\ldots,\lambda_p \in [0,1]$ such that $\sum_{i=1}^p \lambda_i = 1$, we have:
$$\Big ( Choquet(x^1  ,\omega) = \ldots =  Choquet(x^p  ,\omega) \Big )  \Rightarrow   \forall k  \in  \{1, \ldots, p\},  Choquet(x^k ,\omega) \ge  Choquet( \bar{x} , \omega)$$
where $\bar{x}$ is the average solution whose cost vector is $y(\bar{x})=\sum_{i=1}^p \lambda_i y(x^i)$.
\end{prop}

For example, when using a concave capacity in minimization problems, if the DM is indifferent between cost vectors $(0, 1) $ and $(1, 0)$, then we know that she prefers cost vector $(0.5, 0.5)$ to any of the two other vectors, as $(0.5, 0.5)$ is obtained by averaging these two vectors ($\lambda_1 = 0.5$, $\lambda_2 = 0.5$). Hence using a concave capacity enables to favor well-balanced vectors in minimization problems (we obtain the reverse preference when using a convex capacity). 
Obviously, one needs to use a convex (resp. concave) capacity to favor well-balanced (resp. unbalanced) solutions in maximization problems.


Before introducing another useful formulation of the Choquet integral, we now provide an alternative formulation of capacities using M\"obius masses \cite{ChJa89}:

\begin{defn}[M\"obius Inverse and M\"obius Masses] \label{mobiuscapacity}
Any capacity $\omega :  2^N \rightarrow \mathbb{R}$  is associated with a set-function $m :  2^N \rightarrow \mathbb{R}$ called M\"obius inverse, which is defined by
\begin{equation}
\forall A\subseteq N, m(A)=\sum_{B\subseteq A}{{(-1)}^{|A\backslash B|}\omega(B)}, \notag
\end{equation}
so that $\omega$ can be reconstructed from its M\"obius inverse as follows:
\begin{equation}
 \forall A \subseteq N, \omega(A)=\sum_{B\subseteq A}{m(B)}. \notag
\end{equation}
The coefficients $m(A)$ for all $A \subseteq N$ are called M\"obius masses.
\end{defn}

Interestingly, any capacity whose M\"obius masses are non-negative (a.k.a. belief functions) is necessarily convex \cite{Shafer76}. Now using the M\"obius inverse, we can define the notion of $k$-additive capacity \cite{Grabischetal09}. 
A capacity is said to be $k$-additive when its M\"obius masses $m(A)$ are equal to zero for all $A\subseteq N$ such that $|A|>k$, and there exists at least one set $A$ of size $k$ such that $m(A)\neq 0$. More formally:

\begin{defn} [$k$-Additive Capacity]
A capacity is said to be $k$-additive when $m$ the associated M\"obius inverse is such that:
$$\Big (\forall A\subseteq N, |A|>k\Rightarrow m(A)=0 \Big ) \mbox{ and } \Big ( \exists A\subseteq N, |A|=k~\mbox{and}~m(A)\neq 0 \Big )$$
\end{defn} 

Note that $k$-additive capacities correspond to additive capacities when $k=1$. 
For small values of $k$ (greater than $1$), $k$-additive capacities are very useful in practice as they enable to model interactions between criteria/objectives with a reduced number of preference parameters. For example, for the subclass of 2-additive capacities, capacities are completely characterized by $(n^2+n)/2$ values (one M\"obius mass for every singleton and every pair). Using the notion of M\"obius inverse, we obtain the following formulation of the Choquet integral:
\begin{equation}
Choquet(x,\omega) = \sum_{A \subseteq N}  m(A) \min_{j \in A} y_j(x) \label{eq:choquet-moebius}
\end{equation}
This makes explicit another interpretation of the Choquet integral: it is a weighted sum applied on the real-valued vector of size $2^n$ whose components are $\min_{j \in A} y_j(x)$ for $A \subseteq N$.

\subsection{Minimax Regret} 

In this paper, we assume that the DM's preferences are represented by a parameterized aggregation function $f_\omega$ that is linear in its parameters (e.g., a weighted sum, an OWA operator, a Choquet integral), and we assume that the parameters are not known initially. 
Instead, we have a set $\Theta$ of pairs $(u,v)\in \mathbb{R}^n\times \mathbb{R}^n$ such that point $u$ is preferred to point $v$ (which has been obtained by asking preference queries to the DM). Let $\Omega_\Theta$ be the set of all parameters that are compatible with $\Theta$, i.e. all parameters $\omega$ that satisfy the  constraints $f_\omega(u) \le f_\omega(v)$ for all $(u,v) \in \Theta$. Since function $f_\omega$ is linear in its parameters, we can assume that $\Omega_\Theta$ is a convex polyhedron throughout the paper without loss of generality.

\medskip

The problem is now to determine the most promising solution under the preference imprecision (defined by $\Omega_\Theta$). To do so, we use the minimax regret approach (e.g., \cite{boutilier-minimaxAIJ06}) which is based on the following definitions:

\begin{defn}[Pairwise Max Regret] 
The Pairwise Max Regret (PMR) of solution $x \in \mathcal{X}$ with respect to solution $x' \in \mathcal{X}$ is defined as follows:
$$
PMR(x,x',\Omega_\Theta) = \max_{\omega\in \Omega_\Theta} \Big \{ f_\omega(y(x)) - f_\omega(y(x')) \Big \}
$$
\end{defn}
In other words, $PMR(x,x',\Omega_\Theta)$ is the worst-case loss when choosing solution $x$ instead of  $x'$. 

\begin{defn}[Max Regret] \label{defMR}
The Max Regret (MR) of solution $x \in \mathcal{X}$ is defined as follows:
$$
MR(x, \mathcal{X},\Omega_\Theta) = \max_{x' \in \mathcal{X}} PMR(x,x',\Omega_\Theta)
$$
\end{defn}
Thus $MR(x, \mathcal{X},\Omega_\Theta)$ is the worst-case loss when selecting solution $x$ instead of any other feasible solution $x' \in \mathcal{X}$. We can now define the minimax regret:

\begin{defn}[Minimax Regret] \label{defMMR}
The MiniMax Regret (MMR) is defined as follows:
$$
MMR(\mathcal{X},\Omega_\Theta) = \min_{x \in \mathcal{X}} MR(x,\mathcal{X},\Omega_\Theta)
$$
\end{defn}
According to the minimax regret criterion, $x$ is an optimal solution iff $x \in \arg\min_{x \in \mathcal{X}} MR(x,\mathcal{X},\Omega_\Theta)$ (i.e., $MR(x,\mathcal{X},\Omega_\Theta)=MMR(\mathcal{X},\Omega_\Theta)$). Recommending such a solution guarantees that the worst-case loss is minimized (given the imprecision surrounding the DM's preferences).


Note that if $MMR(\mathcal{X},\Omega_\Theta) = 0$, then any optimal solution for the minimax regret criterion is necessarily optimal according to the DM's preferences. 
Note also that we have $MMR(\mathcal{X},\Omega_{\Theta'}) \le MMR(\mathcal{X},\Omega_\Theta)$ for any set $\Theta' \supseteq \Theta$, as already observed in previous works (see e.g., \cite{benabbouAIJ17}). These two observations have led to the following incremental eliciation approach (also called {\em regret-based incremental elicitation}): Progressively collect preference information from the DM by asking her preference queries, one after the other, until $MMR(\mathcal{X},\Omega_\Theta)$ drops below a given tolerance threshold $\delta \ge 0$ (representing the maximum allowable gap to optimality); if we set $\delta = 0$, then we obtain the (optimal) preferred solution at the end of the execution.

\section{A Regret-Based Incremental Genetic Algorithm}

\subsection{Algorithm Description}

To implement the regret-based incremental elicitation approach, one needs to compute the value $MMR(\mathcal{X},\Omega_\Theta)$ at every iteration step of the procedure. For MOCO problems, this may induce prohibitive computation times since it may require to compute the pairwise max regrets for all pairs of distinct solutions in $\mathcal{X}$ (see Definition \ref{defMMR}). Therefore, we propose instead to combine search and regret-based incremental elicitation to reduce both computation times and number of preference queries. More precisely, we now introduce an interactive genetic algorithm that uses regret-based incremental elicitation techniques to select individuals from populations.
 Our algorithm, called RIGA for \emph{Regret-based Incremental Genetic Algorithm}, follows the traditional scheme of genetic algorithms but differs in the following way:
\begin{itemize}
    \item Our populations $P$ are composed of pairs ($\omega,x_\omega$), where $\omega$ is a possible instance of the preference parameters and solution $x_\omega$ is $f_{\omega}$-optimal (or almost).
    \item The crossover and mutation operators are applied on parameter instances not on solutions.
\end{itemize}

\noindent RIGA proceeds as described in Algorithm 1.

\begin{algorithm}[!h]
\caption{: Regret-Based Incremental Genetic Algorithm \small}\label{algoIEGA}
\begin{algorithmic}
\STATE \textbf{IN} $\downarrow$: $\mathcal{P}$: the MOCO problem; $f_\omega$: the scalarizing function with unknown parameters $\omega$;  $\delta$: the tolerance threshold; $\Theta$: the initial set of preference statements; $S$: the population size; $\mu$: the mutation rate; $K$: the number of elements to be selected at each step.
\STATE \textbf{OUT} $\uparrow$: a solution in $\mathcal{X}$. 
\vspace*{0.25\baselineskip}
\STATE  -$\,$-$|$ Initialization of the admissible parameters:
\STATE $\Omega_\Theta \leftarrow \{\omega : \forall (a,b) \in \Theta, f_\omega(a) \le f_\omega(b)\}$
\STATE  -$\,$-$|$ Generation of the initial population: 
\STATE $P \leftarrow $ \texttt{ComputeExtremePoints\&Solutions}$(\Omega_\Theta,\mathcal{P})$
\STATE -$\,$-$|$ Genetic Algorithm: 
\FOR{$1$ to $M$}
\STATE $P \leftarrow  P \ \cup$ \texttt{Crossover\&Mutation}$(P,S, \mu,\mathcal{P})$
\WHILE{$MMR(X_{P},\Omega_\Theta)$ $>$ $\delta$}
\STATE -$\,$-$|$ Ask the DM to compare two solutions:
\STATE $(x,x') \leftarrow$  \texttt{Query}($X_{P}$)
\STATE -$\,$-$|$ Update preference information:
\STATE $\Theta \leftarrow \Theta \cup \{(y(x),y(x'))\}$
\STATE $\Omega_\Theta \leftarrow \{\omega : \forall (a,b)\in \Theta, f_\omega(a) \le f_\omega(b)\}$
\ENDWHILE
\STATE -$\,$-$|$ Move to the next population: 
\STATE  $x^* \leftarrow \arg\min_{x \in X_{P}} MR(x,X_{P},\Omega_\Theta)$ 
\STATE  $P \leftarrow $ \texttt{Selection}$(x^*,P, K)$
\ENDFOR
\RETURN $x^*$
\end{algorithmic}
\end{algorithm}

\paragraph{Initial Population} To generate the initial population, we have to generate a set of possible preference parameters. Then, for every generated parameter $\omega$, we must determine a solution $x_\omega$ that is (near-)optimal for the \emph{precise} scalarizing function $f_{\omega}$. To do so, we use an existing poly-time solving algorithm (e.g., Prim algorithm for the minimum spanning tree problem with a weighted sum). These parameters could be generated uniformly at random, leading to a poly-time initialization phase. However, we propose instead to generate the extreme points of polyhedron $\Omega_\Theta$, as it gives better results in practice. 


\paragraph{Crossover and Mutation} As already mentioned, we perform crossovers and mutations on parameter vectors (not on solutions). For every resulting parameter vector $\omega$, we proceed as follows: we determine a solution $x_\omega$ that is $f_{\omega}$-optimal (or almost) using an existing efficient solving method, and then we add the pair $(\omega,x_\omega)$ in population $P$. 
To obtain a population of the desired size, we create new parameter vectors by means of convex combinations of vectors in $P$. This crossover operator is of particular interest in optimization problems with imprecise preference parameters because it only generates new admissible parameters from admissible parameters. 
In our experiments, we create a new parameter vector $\omega$ from two parameter vectors $\omega',\omega''$ in $P$ as follows: 
$$
\omega = \lambda \omega' + (1-\lambda) \omega''
$$ where $\lambda \in (0,1)$ is generated uniformly at random.
Then, given a mutation rate $\mu$, we perform Gaussian mutations on single objectives. This mutation operator yields very good results in practice, but more sophisticated operators would be worth investigating.

\paragraph{Selection} To create the new generation, we select $K$ promising pairs from population $P$ as follows:
\begin{itemize}
    \item First, we determine a (near-)optimal solution in population $P$ by means of a regret-based incremental elicitation approach. More precisely, let $X_P$ be the set of all solutions in $P$. 
    While $MMR(X_P, \Omega_\Theta) > \delta$, the DM is asked to compare two solutions $x,x'$ and state which one is preferred over the other one. The set $\Omega_\Theta$ of admissible parameters is then updated by inserting the linear constraint $f_\omega(x) \le f_\omega(x')$ or $f_\omega(x) \ge f_\omega(x')$, depending on her answer.  
   In our experiments, we use the Current Solution Strategy (CSS) \cite{boutilier-minimaxAIJ06} to generate the preference queries. This strategy consists in asking to compare a solution $x\in X_P$ achieving the minimax regret to one of its adversary's choice (i.e. a solution in $\arg\max_{x'\in X_P} PMR(x,x',\Omega_\Theta)$) at each iteration step of the elicitation process.
     Once the value $MMR(X_P, \Omega_\Theta)$ drops below threshold $\delta$, we stop asking queries and select a solution $x^*$ that is optimal for the minimax regret criterion, i.e. a solution $x^*$ in $ \arg\min_{x \in X_P} MR(x,X_P,\Omega_\Theta)$.
    \item Then, we compute the distance in objective space between $x^*$ and $x$ for every pair $(\omega,x)$ in $P$ and we select the $K$ pairs that minimize the distance to breed the next generation. 
     In our experiments, we use the Euclidean distance but other distances would be worth investigating.
\end{itemize}

\paragraph{Termination} RIGA stops after $M$ steps and returns a solution arbitrary chosen in  $\arg\min_{x \in X_P}$ $MR(x,X_P,\Omega_\Theta)$, where $P$ is the last generated population.

\subsection{Performance Guarantees}

\medskip

This subsection is devoted to the proof  of the following result:

\begin{prop} 
For any MOCO problem with preferences represented by an OWA operator, if the problem can be solved (exactly or approximately) in polynomial time (in the problem size) when the preferences are precisely known, then RIGA can be implemented in such way that it runs in polynomial time and asks no more than a polynomial number of queries. The same applies to Choquet integrals. 
\end{prop}

Note that our algorithm has different tunable parameters: $S$ the size of the population (generated by crossovers and mutations), $K < S$ the number of pairs selected for the next generation and $M$ the number of steps of the for loop. In order to obtain a poly-time algorithm, we must clearly set $M$ and $S$ to some integer values that are also polynomial.

\subsubsection{OWA operators}

\medskip

\paragraph{On the number of preference queries} 
Note that the number of preference queries generated by RIGA is equal to the number of steps of the while loop. At every step of the for loop, the while loop iterates until the value $MMR(X_P, \Omega_\Theta)$ drops below $\delta$.  We know that this value will be equal to zero after asking the DM to compare all solutions in $X_P$. 
When using the CSS query generation strategy, the DM is asked to compare two solutions in $X_P$ at every step of the while loop. Therefore, RIGA implemented with the CSS strategy generates at most $|X_P|^2$ comparison queries at every step of the for loop. 
Since $M$ (the number of steps of the for loop) and $|X_p|=S$ (the population size) are polynomial, then the number of preferences queries is also polynomial.

\paragraph{On the computation times} We have just proved that the number of iteration steps is polynomial when applying the CSS strategy. Therefore, we only need to prove that MMR-computations can be performed in polynomial time.  Recall that $MMR(X_P,\Omega_\Theta)$ can be obtained by solving at most $|X_P|^2 = |S|^2$ pairwise max regrets (see Definitions \ref{defMR} and \ref{defMMR}). Therefore, it is sufficient to show that pairwise max regrets can be computed in polynomial time. For OWA aggregators, even though $f_\omega(a)=\sum_{j=1}^n \omega_j a_{(j)}$ is non-linear in $a$ for fixed parameters $\omega$, it is linear in $\omega$ for fixed $a$. Thus $PMR(x,x',\Omega_\Theta)$ can be obtained in polynomial time by solving the following linear program:

   \begin{align} 
 \max_{\omega} \;\;\;\; & \sum_{j=1}^n \omega_j y_{(j)}(x') - \sum_{j=1}^n \omega_j y_{(j)}(x) \notag \\
 s.t. \;\;\;\; &  \sum_{j\in N} \omega_j = 1 \notag \\ 
 & \sum_{j=1}^n \omega_j a_{(j)} - \sum_{j=1}^n \omega_j b_{(j)} \le 0, \;\forall \, (a , b) \in \Theta \notag \\
& \omega_{j} \geq 0,\, \forall j \in N \notag 
\end{align}

The constraints $\omega_j \le  \omega_{j+1}$ for all $j \in \{1,\ldots,n-1\}$ (resp.  $\omega_j \ge  \omega_{j+1}$ for all $j \in \{1,\ldots,n-1\}$) can be added so as to favor well-balanced solutions in minimization (resp. maximization) problem.

\subsubsection{Choquet integrals}

As regards the number of queries, the same reasoning as for OWA aggregators applies for Choquet integrals when considering the CSS query generation strategy. This is not the case for computation times. More precisely, since $f_\omega(a)$ is linear in $\omega$ for fixed $a$ with Choquet integrals, $PMR(x,x', \Omega_\Theta)$ can also be obtained by solving the following program:

\begin{figure}[h] 
    \centering
\begin{align} 
  \max_{\omega} & \sum_{j=1}^n \Big ( \big (y_{(j)}(x') - y_{(j-1)}(x') \big ) \omega(X'_{(j)}) - \big (y_{(j)}(x) - y_{(j-1)}(x)\big ) \omega(X_{(j)}) \Big )\notag \\
s.t. \; & \omega(\emptyset) = 0 \notag \\
& \omega(N) =1 \notag \\
& \omega(A) \leq  \omega(B), \; \forall A \subset B \subseteq N \notag\\
& \sum_{j=1}^n \Big ( \big (a_{(j)} - a_{(j-1)}\big ) \omega(A_{(j)}) - \big (b_{(j)} - b_{(j-1)}\big )  \omega(B_{(j)}) \Big ) \le 0, \, \forall (a,b) \in \Theta \notag \\
& \omega(A) \ge 0, \, \forall A \subseteq N \notag
\end{align}
\end{figure}

However, in this linear program, the number of variables and constraints is exponential in the number of variables. This linear program can be simplified for some subclasses of Choquet integrals defined using its alternative formulation in terms of M\"{o}bius masses  (see Equation \ref{eq:choquet-moebius}). For instance, for the subclass of belief functions, the monotonicity constraints $\omega(A) \leq  \omega(B)$, with $A \subset B \subseteq N$, are naturally satisfied due to the non-negativity of the M\"{o}bius masses.  For the subclass of 2-additive capacities, it has been proposed to use the fact that they form a convex polytope with only $n^2$ extreme points (see e.g., \cite{TehraniCDH12}). This yields a linear formulation only involving a quadratic number of variables and constraints. Hence for 2-additive capacities, RIGA combined with the CSS strategy runs in polynomial time and generates no more than a polynomial number of queries.

For general capacities, it has been shown in \cite{BenabbouPV14} that PMR-optimization problems can be formulated as linear programs with only a linear number of variables and constraints, provided that the DM is only asked to compare binary alternatives to constant utility profiles. In the same paper, the authors propose a query generation strategy that is very efficient in practice to reduce the minimax regret values. 
The proposed query generation strategy can be adapted to ensure a polynomial number of queries by simply selecting the constant utility profile in the middle of the interval representing the possible capacity values, instead of selecting the query that minimizes the worst-case minimax regret at each step (as proposed in \cite{benabbouPernyIjcai2017}). Thus, for general Choquet integrals, RIGA can be implemented in such way that it runs in polynomial time and generates no more than a polynomial number of queries.

\subsection{Example}

\medskip

\begin {figure}[!ht]
\begin{center}
\includegraphics [width =8cm]{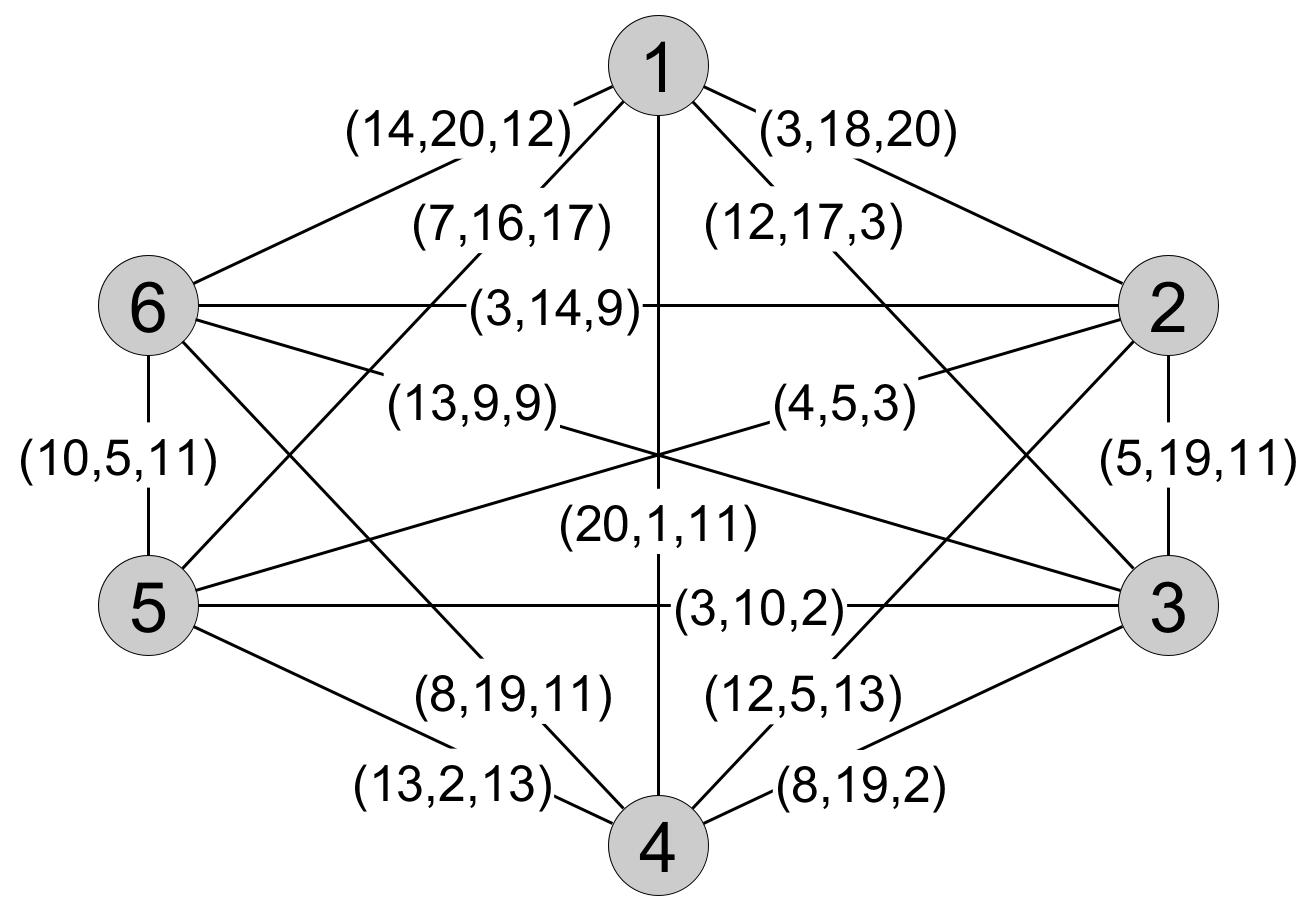}
    \caption {A MTSP instance with 6 vertices and $3$ objectives.  \label{graphExemple}}
\end{center}
\end {figure}

Now we present an execution of RIGA on a small instance of the multi-objective traveling salesman problem (MTSP) with $6$ cities and $n=3$ additive cost functions to be minimized (see Figure \ref{graphExemple}) . In this optimization problem, the set $\mathcal{X}$ of feasible solutions is the set of all Hamiltonian cycles, i.e. cycles that include every node exactly once. We assume here that the DM's preferences over Hamiltonian cycles can be represented by an OWA operator with the hidden increasing weight $\omega^* = (0.1, 0.3, 0.6)$. We now apply RIGA on this instance with $\delta=0$, $S=5$, $K=2$, $M=2$.

\medskip

\medskip

{\bf Initialization phase:}  The set of  admissible weighting vectors is initially defined by  $\Omega_\Theta=\{\omega=(\omega_1,\omega_2,\omega_3) \in [0,1]^3 : \omega_1+\omega_2+\omega_3=1 \mbox{ and } \omega_1 \leq \omega_2 \leq \omega_3\}$. Note that we can assume that $\Omega_\Theta$ is a convex polyhedron throughout this subsection since any constraint of type  $f_\omega(a) \le f_\omega(b)$ is linear in $\omega$ for any fixed performance vectors $a,b$.  In Figure \ref{figInit}, the initial set $\Omega_\Theta$ is represented by the triangle ABC in the space $(\omega_1,\omega_2)$, $\omega_3$ being implicitly defined by $\omega_3 = 1-\omega_1-\omega_2$. 
The extreme points of $\Omega_\Theta$ are $(0,0.5,0.5)$, $(\frac{1}{3},\frac{1}{3},\frac{1}{3})$ and $(0,0,1)$. 

\begin{figure}[!htb]
\minipage{0.32\textwidth}
\begin{tikzpicture}[scale=0.28]\footnotesize
\coordinate (B) at (0,10);
\coordinate (A) at (0,0);
\coordinate (C) at (10,0);
\coordinate (D) at (2, 6);
\coordinate (F) at (6.6,6.6);
\fill[color=green!20!blue!80!white, opacity=0.3] (A) -- (B) -- (F)-- cycle;

\draw (B)--(F);
\draw (F)--(A);

\draw[-latex] (A)-- (0,11.5);
\draw[-latex] (A)-- (11.5,0);
\draw (12.5,0.7) node[below]{$\omega_1$};
\draw (0,11.5) node[above]{$\omega_2$};
\draw (F) node{$\bullet$};
\draw (D) node{$\bullet$};
\draw (2.5, 6.4) node[above]{$\omega^*$};
\draw (B) node{$\bullet$};
\draw (B) node[above right]{$B$};
\draw (B) node[left]{$0.5$};
\draw (A) node{$\bullet$};
\draw (F) node[above right]{$C$};
\draw (A) node[below right]{$A$};
\draw (A) node[below left]{$0$};
\draw (C) node{\tiny $+$};
\draw (C) node[below]{$0.5$};
\end{tikzpicture}
\caption{Initial set $\Omega_\Theta$.}\label{figInit}
\endminipage\hfill
\minipage{0.32\textwidth}
\begin{tikzpicture}[scale=0.28]\footnotesize
\coordinate (B) at (0,10);
\coordinate (A) at (0,0);
\coordinate (C) at (10,0);
\coordinate (D) at (2, 6);
\coordinate (F) at (6.6,6.6);
\coordinate (G) at (1.66, 1.66); 
\coordinate (H) at (0.66, 9.666); 

\fill[color=green!20!blue!80!white, opacity=0.15] (B) -- (H) -- (G)-- (A) -- cycle;
\fill[color=green!20!blue!80!white, opacity=0.30] (G) --(H) -- (F) -- cycle;

\draw (G)--(H);
\draw (G)--(F);
\draw (F)--(H);

\draw[-latex] (A)-- (0,11.5);
\draw[-latex] (A)-- (11.5,0);
\draw (12.5,0.7) node[below]{$\omega_1$};
\draw (0,11.5) node[above]{$\omega_2$};
\draw (F) node{$\bullet$};
\draw (D) node{$\bullet$};
\draw (2.5, 6.4) node[above]{$\omega^*$};
\draw (B) node{\tiny $-$};
\draw (B) node[left]{$0.5$};
\draw (A) node{\tiny $+$};
\draw (F) node[above right]{$C$};
\draw (A) node[below left]{$0$};
\draw (G) node[below right]{$E$};
\draw (H) node[above right]{$D$};
\draw (G) node{$\bullet$};
\draw (H) node{$\bullet$};
\draw (C) node{\tiny $+$};
\draw (C) node[below]{$0.5$};
\end{tikzpicture}
\caption{$\Omega_\Theta$ after $1$ query.}\label{figIter1}
\endminipage\hfill
\minipage{0.32\textwidth}%
\begin{tikzpicture}[scale=0.28]\footnotesize
\coordinate (B) at (0,10);
\coordinate (A) at (0,0);
\coordinate (C) at (10,0);
\coordinate (D) at (2, 6);
\coordinate (F) at (6.6,6.6);
\coordinate (H) at  (0.66, 9.666);  
\coordinate (I) at (2.5, 2.5); 
\coordinate (J) at (1.382, 3.936); 

\fill[color=green!20!blue!80!white, opacity=0.3] (H) -- (F) -- (I)-- (J) --  cycle;
\fill[color=green!20!blue!80!white, opacity=0.15] (B) -- (A) -- (I)-- (J) -- (H) -- cycle;
\draw (H)--(F);
\draw (F)--(I);
\draw (I)--(J);
\draw (J)--(H);

\draw[-latex] (A)-- (0,11.5);
\draw[-latex] (A)-- (11.5,0);
\draw (12.5,0.7) node[below]{$\omega_1$};
\draw (0,11.5) node[above]{$\omega_2$};
\draw (F) node{$\bullet$};
\draw (D) node{$\bullet$};
\draw (2.5, 6.4) node[above]{$\omega^*$};
\draw (B) node{\tiny $-$};
\draw (B) node[left]{$0.5$};
\draw (A) node{\tiny $+$};
\draw (F) node[above right]{$C$};
\draw (A) node[below left]{$0$};
\draw (I) node[below right]{$F$};
\draw (1.6, 4) node[below left]{$G$};
\draw (H) node[above right]{$D$};
\draw (I) node{$\bullet$};
\draw (J) node{$\bullet$};
\draw (H) node{$\bullet$};
\draw (C) node{\tiny $+$};
\draw (C) node[below]{$0.5$};
\end{tikzpicture}
\caption{$\Omega_\Theta$ after $2$ queries.}\label{figIter2}
\endminipage
\end{figure}

\medskip
\begin{sloppypar}
{\bf Initial population:} In order to generate the initial population, we determine one near-optimal solution for each extreme point of polyhedron $\Omega_\Theta$ (using a local search procedure for example). We obtain $P= \{(\omega^A,x^A),(\omega^B,x^B),(\omega^C,x^C)\}$ where $\omega^A=(0,0.5,0.5)$, $\omega^B=(\frac{1}{3},\frac{1}{3},\frac{1}{3})$, $\omega^C=(0,0,1)$, $y(x^A)=(49,52,60)$, $y(x^B)=(39,50,66)$, and $y(x^C)=(56,57,58)$.
\end{sloppypar}
\medskip

{\bf First iteration step:} Since $|P|=3$ and $S = 5$, we need to generate $2$ more pairs. Applying the crossover operator on $\omega^A=(0,0.5,0.5)$ and $\omega^B=(\frac{1}{3},\frac{1}{3},\frac{1}{3})$, and then performing a Gaussian mutation on the first objective, we obtain the following vector $\omega^1= (0.27,0.33,0.40)$. Function $f_{\omega^1}$ is then optimized, resulting in the generation of solution $x^1$ whose cost vector is $(39,50,66)$. 
The pair $(\omega^1,x^1)$ is then inserted in $P$. When applying the crossover operator on $\omega^B=(\frac{1}{3},\frac{1}{3},\frac{1}{3})$ and $\omega^C=(0,0,1)$, and after performing a Gaussian mutation on the second objective, we obtain $\omega^2= (0.27,0.33,0.40)$ and cycle $x^2$ whose cost vector is $(56,57,58)$. The pair $(\omega^2,x^2)$ is then inserted into $P$. Now we have a complete population.

The selection stage begins. We ask queries to the DM until $MMR(X_P, \Omega_\Theta) \le \delta=0$, where $X_P=(x^A,x^B,x^C,x^1, x^2)$. Since $MMR(X_P,\Omega_\Theta) = 2 > 0$, we ask the DM to compare two solutions in $X_P$, say $x^A$ and $x^B$. 
Since $f_{\omega^*}(y(x^A)) = 56.5 < f_{\omega^*}(y(x^B)) = 58.5$, the DM answers: ``solution $x^A$ is better than solution $x^B$''. 
Then $\Theta$ (the set of preference statements) is updated by adding the pair $(y(x^A),y(x^B))$; thus we have $\Theta=\{((49, 52, 60), (39, 50, 66))\}$. 
Consequently, the set of admissible parameters $\Omega_\Theta$ is restricted by the linear constraint $f_\omega(y(x^{A})) \le f_\omega(y(x^{B}))$, i.e. $\omega_2 \le \frac{3}{4}-2 \omega_1$.
Now $\Omega_\Theta$ is represented by the triangle DCE in Figure \ref{figIter1}, and we have $MMR(X_P,\Omega_\Theta)= 2 > 0$. Since the minimax regret is above the threshold, we ask the DM to compare two solutions in $X_P$, say $x^C$ and $x^B$. The DM prefers solution $x^B$ to solution $x^C$ since we have $f_{\omega^*}(y(x^B)) = 56.5 < f_{\omega^*}(y(x^C)) = 57.5$. 
We then update $\Theta$ by inserting the pair $((49, 52, 60), (56, 57, 58))$ and we restrict $\Omega_\Theta$ by imposing the linear constraint $f_\omega(y(x^A))\le f_\omega(y(x^C))$, i.e. $\omega_2 \le \frac{2}{7}-\frac{9}{7} \omega_1$ ($\Omega_\Theta$ is now represented by DCFG in Figure \ref{figIter2}). Now we have $MMR(X_P, \Omega_\Theta) = MR(x^A,X_P, \Omega_\Theta) = 0$ (the while loop stops). 
We must select solutions for the next population. Here we have $x^*=x^A$. Since $K=2$, we need to select one more pair for the next generation.  
We choose $x^C$ as it is the closest solution to $x^*$ according to the Euclidean distance. Thus we have $P=\{(\omega^C,x^C), (\omega^A,x^A)\}$ for the next iteration step.

\medskip
\begin{sloppypar}

{\bf Second iteration step:} Since $|P| = 2$ and $S=5$, we have to generate three more pairs. After applying the crossover and mutation operators on  $\omega^A$ and $\omega^C$, we obtain $(0,0.41,0.59)$,  $(0,0.21,0.79)$ and $(0,0.18,0.82)$. We then optimize the corresponding OWA functions and we obtain solutions $x^3$, $x^4$ and $x^5$ whose cost vectors are $y(x^3)=(49,52,60)$, $y(x^4)=(56,57,58)$ and $y(x^5)= (56,57,58)$ respectively. 
Therefore we have $P=\{(\omega^A,x^A), (\omega^C,x^C), (\omega^3,x^3),(\omega^4,x^4), (\omega^5,x^5),(\omega^5,x^5)\}$. 
We do not need to ask a query at this step because $MMR(X_P,\Omega_\Theta) = MR(x^A,X_P, \Omega_\Theta) = 0 \le \delta$. At this step, we have $x^* = x^A$. 
\end{sloppypar}

\medskip

Since $M=2$, RIGA only performs two iteration steps and then stops by returning solution $x^* = x^A$ (corresponding to cycle $2-0-3-1-5-4-2$),  which is here optimal according to the DM's preferences. In this example, only $2$ queries were needed to discriminate between the $17$ Pareto-optimal solutions.

\section{Experimental Results}

In this section, we provide numerical results aiming to evaluate the performances of our algorithm in terms of calculation times (in seconds), number of queries and gap to optimality (empirical error), expressed as a percentage of the optimal value. 
In our experiments, we consider two MOCO problems: the multi-objective knapsack problem (MKP) and the multi-objective traveling salesman problem (MTSP). Numerical tests were performed on a Intel Core i7-8550U CPU with 16 GB of RAM, with a program written in C\texttt{++}, and linear programs are solved using CPLEX Optimizer\footnote{\url{https://www.ibm.com/analytics/cplex-optimizer}}. All the results have been obtained by averaging over 50 runs. 


\subsection{Preferences}

In our experiments, we assume that the DM's preferences are represented by a parameterized aggregation function $f_\omega$ that is linear in its parameters $\omega$. Here we focus on three families of aggregation functions: weighted sums, OWA operators with monotonic weights, and Choquet integrals with $2$-additive capacities.  
Initially, the set of collected preference information is empty (i.e. $\Theta = \emptyset$). 
The answers to queries are simulated using a weighting vector $\omega^*$ randomly generated before running the methods. We use the process described in ~\cite{RUBINSTEIN1982205} to ensure an even distribution of weights.





\subsection{Implemented Methods}

We have implemented three versions of our algorithm: 
\begin{itemize}
    \item RIGA: the proposed algorithm. Recall that our algorithm uses an existing solving method to generate solutions from a given weighting vector. For the MKP, we use a simple greedy algorithm for the weighted sum model, while linearization methods are used for OWA \cite{OGRYCZAK200380} and for Choquet integrals \cite{LESCA2013}. For the MTSP, we use the exact TSP solver Concorde\footnote{\url{http://www.math.uwaterloo.ca/tsp/concorde}} for the weighted sum model. For OWA and Choquet integrals, we use the same linearization methods as before implemented with a symmetric TSP formulation and a branch-and-cut. 
    \item RIGA$_{KCSS}$: the proposed algorithm but instead of using the Euclidean distance to select solutions at each iteration step, we select solutions one by one using the CSS query generation strategy. More precisely, we first generate preference queries according to the CSS strategy, until identifying a solution with a max regret below the given tolerance threshold. Then, this solution is removed from the population, and the selection process is iterated on the remaining set of solutions with the same tolerance threshold. This process stops after selecting $K$ solutions. 

    \item RIGA$_S$: the proposed algorithm but instead of applying genetic operators on parameter vectors, they are directly applied on solutions. Here we use the same genetic operators as that of NEMO, i.e. one-point crossovers and swap mutations. 
\end{itemize}

\medskip

We compare our performances to those obtained by the following existing methods:

\paragraph{Interactive Local Search (ILS) \cite{BenabbouLLP19}} 
This interactive local search procedure starts from a promising solution generated using a heuristic method. For the MKP problem, we use the heuristic method proposed in \cite{benabbouLL20} designed for the MKP, which consists in determining an optimal solution according to the arithmetic mean using a greedy algorithm. For the MTSP, we use the general heuristic method proposed in \cite{BenabbouLLP19}, which proceeds as follows: First, a set of possible starting solutions is obtained by optimizing $f_\omega$ using a heuristic, for $100$ randomly generated weighting vectors $\omega$. The heuristic consists in applying a simple local search based on the 2-opt neighborhood~\cite{croes1958method}. Then, preference queries are asked to discriminate between the possible starting points. More precisely, we ask queries until the minimax regret drops below a given threshold $\delta = 0.1$ ($10\%$ of the initial regret), and then the starting solution is arbitrarily chosen among the optimal solutions according to the minimax regret decision criterion.  

Then, we move from solution to solution using a neighborhood function depending on the problem under consideration. More precisely, for the MKP problem, the neighborhood function consists in replacing one item of the current solution by another item (all combinations are performed). 
For the MTSP problem, the neighborhood function is defined by 2-opt swaps. 

In order to select the next solution, preference queries are asked to discriminate between Pareto non-dominated solutions within neighborhoods. More precisely, at every iteration step, we ask queries until the minimax regret drops below a given threshold $\delta = 0.4$ (as proposed in \cite{BenabbouLLP19}), and then we move to a neighbor solution that minimizes the max regret.

\paragraph{Necessary-preference-enhanced Evolutionary Multi-objective Optimizer (NEMO) \cite{Branke2005}} 

In this genetic algorithm, the mutation and crossover operators are applied on solutions (instead of weighting vectors), and a tournament selection method is used to construct populations.  The mutation operator simply consists in exchanging the position of two genes randomly chosen. 
The crossover operator depends on the problem under consideration: we use one-point crossovers for the MKP and two-point crossovers for the MTSP. For the latter problem, we run one local search for every generated solution in order to obtain a population of better quality. More precisely, we start form the generated solution, and we use a simple local search based on the 2-opt neighborhood function. The local improvements are defined by a given aggregation function $f_\omega$, randomly generated before running the local search.   

 At every iteration step, linear programming is used to rank the solutions in the current population, using the collected preference information and the crowding distance. 
In this method, one preference query is generated every 10 generations: the DM is asked to compare two potentially good solutions selected among those in the current population. 
In our experiments, the population size is set to $30$ and the mutation rate is set to $\frac{1}{n}$, as proposed in \cite{Branke2005}. 
Moreover, the number of solutions selected to build the next generation is set to $5$ at every iteration step. The best results were obtained with these parameter values. 

\paragraph{A Two-Phase Method (Two-Phase)} This method consists in first constructing the Pareto set (or an approximation of this set), and then applying the CSS strategy on this set until the minimax regret drops below threshold $\delta \ge 0$. In order to generate the Pareto set efficiently, we use dynamic programming~\cite{BazganHV09} and fast Pareto dominance checking~\cite{JaszkiewiczL18} for the MKP problem. For the MTSP problem, a good-quality approximation is generated by using the heuristic method proposed in \cite{Jaszkiewicz18} which is based on local search. 
We then select randomly $3000$ solutions, to reduce computation times of the CSS strategy. 
Note that the reported calculation times do not include the time required to generate the (approximate) Pareto set. 

\paragraph{Incremental Elicitation based on Extreme Points (IEEP) \cite{BenabbouL19}} The exact solving method introduced in \cite{BenabbouL19} for general MOCO problems, which is based on the extreme points of $\Omega_\Theta$ (the polyhedron representing the admissible preference parameters). 
For every extreme point, an exact solving method is used to obtain the corresponding optimal solution. In our experiments, we use the same solving methods as those used to generate solutions in RIGA. This interactive method is guaranteed to return a solution with a max regret below a given threshold $\delta \ge 0$ at the end of the execution, provided that exact solving methods are used (not heuristics).

\subsection{Multi-Objective Knapsack Problem}

In these experiments, we consider instances of the MKP with $100$ items, and $n=3$ to $6$ objectives.  Instances are generated as follows: performance vectors attached to items are uniformly drawn in $\{1,\ldots, 1000\}^n$ and the knapsack capacity is set to $50$ (half of the number of items) so as to obtain difficult instances (i.e., with a large number of Pareto-optimal solutions).

\subsubsection{RIGA: parameter setting and evaluation}

In order to evaluate the performances of our algorithm, we use different parameter values: $M$ the number of iterations is set to $5,10, 20$, $S$ the size of the population is set to $10, 20, 30$, and $K$ the number of selected pairs is set to $2, 5, 10$. For every generated instance of the MKP problem, we run our algorithm with all possible combinations of parameter values, and the average results are given in Table~\ref{tabParam}.

\begin{table*}[h!]
\centering{\footnotesize
\scalebox{0.9}{

    \begin{tabular}{  c | c | c | rrr|rrr| rrr } 
    \cline{1-12} \multicolumn{2}{c}{} &  &  \multicolumn{3}{c|}{\textbf{WS}} &   \multicolumn{3}{c|}{\textbf{OWA}} & \multicolumn{3}{c}{\textbf{Choquet}}   \\  

     \multicolumn{1}{c}{\bm{$M$}} & \multicolumn{1}{c}{\bm{$S$}} & \bm{$K$} & \textbf{time} & \textbf{queries} & \textbf{error} & \textbf{time} & \textbf{queries} & \textbf{error}  &  \textbf{time} & \textbf{queries} & \textbf{error}  \\ \hline

        \multirow{9}{*}{$5$}
            & \multirow{3}{*}{$10$} & 2 & 0.44    & 22.12   & 1.05  &  5.13    & 5.32    & 0.002 &    6.18    &  39.10   &  0.55      \\ 
            & & 5 & 0.60    & 25.42   & 0.76  &  4.23   & 5.94     & 0.001 &    4.60    &  42.30   &  0.43           \\ 
            & & 10 & 0.65   & 26.12   & 0.85  &  6.65    & 5.40     & 0.003 &    4.45    &  43.47   &  0.56           \\ \cline{2-12}

            & \multirow{3}{*}{$20$} & 2 & 0.72    & 29.34   & 0.45  &  7.65    & 6.48    & 0.001  &    6.94    &  50.03   &  0.33        \\ 
            & & 5 & 0.87    & 29.46   & 0.45  &  7.61    & 6.68     & 0.000 &    7.59    &  53.23   &  0.34           \\ 
            & & 10 & 1.17    & 32.28   & 0.46  &  10.25    & 7.36     & 0.001 &    10.95    &  52.15   &  0.32           \\ \cline{2-12}
      
            & \multirow{3}{*}{$30$} & 2 & 1.43   & 32.36   & 0.29  &  11.93    & 6.74    & 0.000 &    10.94    &  52.15   &  0.31        \\ 
            & & 5 & 1.56    & 35.38   & 0.38  &  11.99    & 7.64     & 0.000 &    11.21    &  57.08   &  0.34           \\ 
            & & 10 & 2.02    & 36.16   & 0.27  &  12.87    & 7.80     & 0.000 &    11.38    &  58.90   &  0.34           \\  \hline

        \multirow{9}{*}{$10$}
            & \multirow{3}{*}{$10$} & 2 & 0.61    & 36.99   & 0.23  &  8.55    & 5.94    & 0.002 &    3.82    &  38.70   &  0.52      \\  
            & & 5 & 0.71    & 33.58   & 0.32  &  9.98    & 6.20     & 0.001 &    4.59    &  42.30   &  0.43           \\ 
            & & 10 & 0.86   & 36.78   & 0.40  &  13.38    & 5.94     & 0.000 &    4.46    &  43.48   &  0.56           \\ \cline{2-12}

            & \multirow{3}{*}{$20$} & 2 & 1.05    & 40.50   & 0.10  &  16.03    & 7.16    & 0.001 &    7.34    &  46.72   &  0.35        \\  
            & & 5 & 1.28    & 42.94   & 0.09  &  14.40    & 6.82     & 0.001 &     7.60    &  50.02   &  0.33           \\ 
            & & 10 & 1.59    & 42.30   & 0.14  &  19.79    & 7.56     & 0.000 &    17.59    &  53.23   &  0.35            \\ \cline{2-12}
      
            & \multirow{3}{*}{$30$} & 2 & 1.32   & 41.04   & 0.08  &  25.14    & 7.14    & 0.003 &    10.95    &  52.15   &  0.32        \\ 
            & & 5 & 1.67    & 45.38   & 0.07  &  23.17    & 8.44     & 0.001 &    11.21    &  57.08   &  0.34           \\ 
            & & 10 & 1.99    & 44.34   & 0.07  &  29.34    & 8.96     & 0.000 &    11.38    &  58.90   &  0.34           \\ \hline

        \multirow{9}{*}{$20$}
            & \multirow{3}{*}{$10$} & 2 & 0.76    & 42.38   & 0.12  &  24.96    & 6.46    & 0.001 &    6.88    &  55.93   &  0.36        \\ 
            & & 5 & 0.84   & 42.96   & 0.12  & 25.90    & 6.60     & 0.000  &    6.96    &  57.88   &  0.38           \\ 
            & & 10 & 0.88  & 46.46   & 0.19  & 24.14    & 7.84     & 0.000 &    7.52    &  61.80   &  0.32           \\ \cline{2-12}

            & \multirow{3}{*}{$20$} & 2 & 1.03    & 46.52   & 0.06  &  31.58    & 7.70    & 0.001 &    17.28    &  61.78   &  0.34        \\ 
            & & 5 & 1.18    & 47.12   & 0.09  &  30.64    & 8.86     & 0.000 &    17.24    &  69.62   &  0.23           \\ 
            & & 10 & 1.62    & 53.02   & 0.08  & 40.41     &     7.32 & 0.000 &    15.32    &  69.18   &  0.30           \\ \cline{2-12}
      
            & \multirow{3}{*}{$30$} & 2 & 1.68    & 49.56   & 0.03  &  67.05    & 8.48 & 0.000 &    23.13    &  67.80   &  0.26        \\ 
            & & 5 & 1.93    & 51.84   & 0.03  &  50.42    & 7.68     & 0.000 &    24.49    &  73.01   &  0.29           \\ 
            & & 10 & 2.19    & 53.72   & 0.05  & 53.31    & 10.04     & 0.000 &   31.13    &  78.72   &  0.29           \\  \hline

    \end{tabular}
    }
}
\vspace{-0.2cm}
\caption{\footnotesize Results obtained by $RIGA$ with $n=5$, $\delta=0$, and $\mu = 0.5$. 
\label{tabParam}}
\end{table*}

In this table, we observe that setting $M=10$, $S=20$, and $K=5$ allows to obtain the best compromise in terms of computation times, number of queries and error. In particular, the error is too high when considering less iteration steps (i.e. when $M=5$), while the number of queries and the computation times are prohibitive when considering more iteration steps (i.e. when $M=20$). Therefore, we will use $M=10$, $S=20$, and $K=5$ in the remaining of this subsection.




\medskip

\begin{table*}[h!]
\centering{\footnotesize
\scalebox{0.9}{

    \begin{tabular}{  c | rrr|rrr| rrr } 
    \cline{1-10} \multicolumn{1}{c|}{}  &  \multicolumn{3}{c|}{\textbf{WS}} &   \multicolumn{3}{c|}{\textbf{OWA}} & \multicolumn{3}{c}{\textbf{Choquet}}   \\  

      \bm{$\mu$}  & \textbf{time} & \textbf{queries} & \textbf{error} & \textbf{time} & \textbf{queries} & \textbf{error}  &  \textbf{time} & \textbf{queries} & \textbf{error}  \\ \hline

        \multirow{1}{*}{$0.2$}
             &   0.95  &  34.92  & 0.33 &  14.37  & 7.00 & 0.00  &   6.31  & 44.20   & 0.64\\ 
            
        \multirow{1}{*}{$0.5$}
             &   1.28   & 42.94 & 0.09   & 14.40   & 6.82  & 0.00   &  7.60  &    50.02 & 0.33\\ 
      
        \multirow{1}{*}{$0.8$}
             &  1.47   & 45.08   & 0.13  & 16.74   & 8.08  & 0.00   & 8.19   & 54.60   & 0.30\\  \cline{1-10}
        
    \end{tabular}
    }
}
\vspace{-0.2cm}
\caption{\footnotesize Results obtained by RIGA with $\delta = 0$ and $n=5$.
\label{tabmut}}
\end{table*}




            
      
        

Now, we aim to evaluate the impact of the mutation rate on the performances of our algorithm. To do so, we use different mutation rates: $\mu=0.2,0.5, 0.8$. The results are given in Table \ref{tabmut}. We observe that the setting $\mu=0.5$ leads to a good compromise between the number of queries, the computation time and the error. In particular, the error increases when considering a smaller value (i.e. $\mu=0.2$) as it does not allow for enough diversity, while the number of queries and the computation times increase when using a larger value (i.e. $\mu=0.8$). Thus we will use $\mu=0.5$ hereafter.

Finally, we consider different tolerance thresholds: $\delta=0,0.3,0.5,0.7$ (see Table \ref{tabdelta}). We observe that setting $\delta=0.5$ leads to a good compromise between the number of queries, the computation time, and the error. For example, using a smaller value induces a greater number of preference queries as more questions are needed to drop below the targeted regret value. Note that increasing $\delta$ reduces the computation times for both WS and Choquet, but not for OWA. This is due to the fact that the method used to solved the problem with known weights is less efficient when considering unbalanced weights, which are more likely to be selected when considering a larger tolerance threshold. 

\begin{table*}[h!]
\centering{\footnotesize
\scalebox{0.9}{

    \begin{tabular}{  c | rrr|rrr| rrr } 
    \cline{1-10} \multicolumn{1}{c|}{}  &  \multicolumn{3}{c|}{\textbf{WS}} &   \multicolumn{3}{c|}{\textbf{OWA}} & \multicolumn{3}{c}{\textbf{Choquet}}   \\  

      \bm{$\delta$}  & \textbf{time} & \textbf{queries} & \textbf{error} & \textbf{time} & \textbf{queries} & \textbf{error}  &  \textbf{time} & \textbf{queries} & \textbf{error}  \\ \hline

        \multirow{1}{*}{$0$}
             &   1.28  &  42.94  & 0.09 &  14.40  & 6.82 & 0.00  &   6.94  & 50.02   & 0.33\\ 
            
      
                    
         \multirow{1}{*}{$0.3$}
              &   0.37   & 13.90   & 0.21  & 15.92 & 5.48  & 0.00   &  6.34  & 14.80   & 0.46\\ 


         \multirow{1}{*}{$0.5$}
              &    0.32  & 13.08   & 0.44  &  17.33  &  4.60 &  0.00  &  6.05  & 12.83   &0.49\\ 


         \multirow{1}{*}{$0.7$}
              &    0.33  & 12.54   & 0.57  &  19.24  &  4.70 &  0.01  &  5.81  & 10.90   &0.45\\ 
            
             \cline{1-10}
        
    \end{tabular}
    }
}
\vspace{-0.2cm}
\caption{\footnotesize Results obtained by RIGA with $n=5$.
\label{tabdelta}}
\end{table*}

\subsection{Possible variants}

In this subsection, we aim to compare the performances of our algorithm to those obtained by its two variants: RIGA$_{KCSS}$ and RIGA$_S$. 
The results are given in Table \ref{tabComkcssS}. 
First, we observe that RIGA$_{KCSS}$ generates a prohibitive number of preference queries, without significantly reducing the error. For example, with $n=5$ and Choquet integrals, $134$ queries are needed against $50$ for RIGA. 
Moreover, RIGA is much faster than RIGA$_{KCSS}$. For instance, RIGA is about $7$ times faster than RIGA$_{KCSS}$ for the weighted sum model and $n=5$ criteria.

Then, we observe that applying genetic operators on parameter vectors is more efficient than applying them on feasible solutions. We indeed observe that RIGA$_S$ asks more queries than RIGA, while obtaining much higher errors. 
Note that the error achieved by RIGA$_S$ could be reduced by increasing $M$ the number of iteration steps (generations), but this would further increase the number of queries. 
Note also that RIGA$_S$ is much faster than RIGA, as it does not require to solve the MKP problem with precise weights at every iteration step.

\begin{table*}[h!]
\centering{\footnotesize
\scalebox{0.9}{

    \begin{tabular}{  c  c | rrr|rrr| rrr } 
    \cline{1-11} \multicolumn{2}{c|}{}  &  \multicolumn{3}{c|}{\textbf{RIGA}} &   \multicolumn{3}{c|}{\textbf{RIGA$_{KCSS}$}} & \multicolumn{3}{c}{\textbf{RIGA$_{S}$}}   \\  

      \multicolumn{1}{c}{\textbf{function}}  & \bm{$n$} & \textbf{time} & \textbf{queries} & \textbf{error} & \textbf{time} & \textbf{queries} & \textbf{error}  &  \textbf{time} & \textbf{queries} & \textbf{error}  \\ \hline

        \multirow{2}{*}{WS}
             & 3 &  0.40  & 16.40   & 0.02 &  5.11  & 24.86     & 0.01 &    0.40    & 14.04    &    5.27        \\ 
             & 5 &  1.28  &  42.94  & 0.09 &  8.40  & 73.72     & 0.04 &   1.57     & 33.20    &  5.39           \\ \cline{1-11}

        \multirow{2}{*}{OWA}
             & 3 &  5.55  & 5.10   & 0.00   & 12.05   & 7.12     & 0.00 &   0.13     & 7.30 & 3.32            \\ 
             & 5 &  14.40  & 6.82 & 0.00 &  28.85  & 10.74     & 0.00 &  0.45  &  12.23  & 3.87           \\ \cline{1-11}
      
        \multirow{2}{*}{Choquet}
             & 3 &  7.03  &  24.14  & 0.64  &  6.36  &71.26  & 0.64  &  0.47      &  31.60   &3.93         \\ 
             & 5 &  7.60  & 50.02   & 0.33  &  12.38  & 134.66  & 0.33  &   2.55     & 64.22    & 3.99            \\ \cline{1-11}

    \end{tabular}
    }
}
\vspace{-0.2cm}
\caption{\footnotesize Results obtained by RIGA, RIGA$_{KCSS}$ and RIGA$_S$ with $\delta = 0$. 
\label{tabComkcssS}}
\end{table*}

\subsubsection{Comparisons to other methods}

In this subsection, we compare the performances of RIGA (with $\delta=0.5$) to those obtained by the methods presented in Section 4. The results are given in Table \ref{tabComp}. For the two exact methods (namely IEEP$_\delta$ and Two-Phase$_\delta$), we set $\delta=0.01$ to allow for the same error as our method.

\begin{table*}[h]
\centering{\footnotesize
\scalebox{0.9}{

    \begin{tabular}{  c  c | rrr|rrr| rrr } 
    \cline{1-11} 
   \multicolumn{2}{c|}{}  &  \multicolumn{3}{c|}{\textbf{WS}} &   \multicolumn{3}{c|}{\textbf{OWA}} & \multicolumn{3}{c}{\textbf{Choquet}}   \\  
     \textbf{method} & \bm{$n$} & \textbf{time} & \textbf{queries} & \textbf{error} & \textbf{time} & \textbf{queries} & \textbf{error}  &  \textbf{time} & \textbf{queries} & \textbf{error}  \\ \hline

        \multirow{4}{*}{RIGA}
             & 3 &  0.18  & 10.4   & 0.14 & 6.70  &  3.9  & 0.003  &  7.24   & 11.0  & 0.68      \\
             & 4 &  0.24  & 12.0   & 0.21 & 11.18 &  4.3  & 0.001  &  6.39   & 12.1  & 0.31        \\ 
             & 5 &  0.32  & 13.1   & 0.44 & 17.33 &  4.6  & 0.001  &  6.05   & 12.8  & 0.49         \\ 
             & 6 &  0.38  & 14.1   & 0.71 & 28.45 &  4.5  & 0.003  &  7.94   & 14.3  & 0.49    \\ \cline{1-11}



             
        \multirow{4}{*}{NEMO}
             & 3 &  13.19  &  15.0 &  0.25 & 18.03  & 15.0  & 0.10  &  10.46  &  15.0   & 0.21          \\
             & 4 &   9.45  &  15.0 & 0.93  & 13.99  & 15.0  & 0.22  &  5.35   &  15.0   & 0.56             \\ 
             & 5 &  7.34   &  15.0 & 1.37  & 11.73  & 15.0  & 0.37  &  5.38   &  15.0   & 0.85             \\ 
             & 6 &  5.60   &  15.0 & 1.86  & 10.63  & 15.0  & 0.40  &  5.17   &  15.0   & 1.29              \\ \cline{1-11}      
        
        \multirow{4}{*}{IEEP$_\delta$}
             & 3 &  11.43 & 8.8   & 0.06  &   7.92  &  6.4   & 0.00 &  54.11    &  23.2  & 0.05            \\
             & 4 &  16.88 & 13.8  & 0.06  &   9.32  &  7.9   & 0.00 &     / & /   & /            \\ 
             & 5 &  22.46 & 18.1  & 0.10  &   29.03 &  19.2  & 0.00 &     / & /   & /           \\ 
             & 6 &  27.46 & 25.1  & 0.19  &   24.99 &  13.5  & 0.05 &     / & /   & /           \\ \cline{1-11}
                    
        \multirow{4}{*}{Two-Phase$_\delta$}
             & 3 & 119.70   &  9.9   & 0.21  &  47.44   & 3.4 & 0.10  & 262.79     & 20.6   & 0.14            \\
             & 4 & 225.02   &  17.4  & 0.12  &  56.54   & 4.2 & 0.26  & 901.03     & 61.9   &    0.10         \\ 
             & 5 & 280.62   &  25.7  & 0.14  &  82.55   & 5.2 & 0.71 &   / & /   & /             \\ 
             & 6 & 372.18   & 32.7   & 0.19  &  73.90   & 4.5 & 1.15 &    / & /   & /           \\ \cline{1-11}
                    
        \multirow{4}{*}{ILS}
             & 3 & 21.83   &  12.6  & 0.05   &  0.32 &  4.0   & 0.08 &  25.23   & 15.6 & 0.89              \\
             & 4 & 81.42   &  22.9  & 0.14   &  0.51 &  4.3   & 0.24 &  108.30 & 25.0 & 0.06             \\ 
             & 5 & 273.67  &  34.0  & 0.08   &  2.39 &  12.5  & 0.25 &  516.40  & 38.8 & 0.11             \\ 
             & 6 & 600.21  &  52.7  & 0.09   &  1.58 &  11.1  & 0.37 &  1056.98 & 49.2 & 0.10         \\ \cline{1-11}
        
    \end{tabular}
    }
}
\vspace{-0.2cm}
\caption{\footnotesize Results obtained by RIGA, NEMO, IEEP$_\delta$, Two-Phase$_\delta$ and ILS. 
\label{tabComp}}
\end{table*}

\paragraph{Results obtained with the weighted sum model} 
We observe that RIGA is much faster than all other methods, whereas Two-Phase$_\delta$ and ILS are too slow. For example, for $n=6$, RIGA is about 1600 times faster than ILS and is almost $1000$ times faster than Two-Phase$_\delta$. 
This is mainly due to the fact that the number of Pareto non-dominated solutions drastically increases with the number of criteria. 
Although these two methods are a little closer to the optimum than RIGA, they ask a greater number of queries than the proposed algorithm. For example, with $n=5$ criteria, RIGA asks $14$ queries against $32$ for Two-Phase$_{\delta}$ and $53$ for ILS. 

When comparing RIGA to IEEP$_{\delta}$, we observe that the latter algorithm obtain smaller errors than the former, but this is at the cost of generating a significantly higher preference queries and requiring larger computation times. To give an example, for $n=6$, the computation time is reduced by a factor of $70$ when using RIGA instead of IEEP$_{\delta}$, and the number of queries is divided by almost $2$. 

Finally, we observe that RIGA is better than NEMO in all the settings. More precisely, when NEMO is stopped after 15 queries (150 generations), RIGA is much faster than NEMO, yields much smaller errors and asks less preference queries.
Note that the execution time of NEMO decreases as the number of criteria increases. 
This can be explained by the fact that constructing the ranking over solutions becomes more and more easier when increasing the number of criteria, as the number of non-dominated solutions increases. 

\paragraph{Results obtained with the OWA model}
First, we observe that the two parameter-based methods, namely RIGA and IEEP$_{\delta}$, obtain very good results, especially on the quality of the returned solution. However, IEEP$_{\delta}$ needs to ask much more queries to provide the desired performance guarantees. 

Then, we observe that ILS is the faster algorithm, but asks too many queries compared to RIGA while generating much larger errors. For example, for $n=5$, ILS is about $7$ times faster than RIGA, but ILS asks about $3$ times more queries, and its error is $250$ times larger.     

When comparing RIGA and Two-Phase$_\delta$, we see that the former outperforms the latter in terms of errors and computation times, while generating similar numbers of queries. 

Finally, we observe that the error achieved by NEMO is surprisingly high when compared to that of the other methods, considering the fact that it was allowed to ask more preference queries.


\paragraph{Results obtained with the Choquet integral model} In the table, we see that the timeout (30 minutes) is exceeded when running IEEP$_\delta$ with 4 criteria and more. 
This can be explained by the fact that the number of extreme points of the polyhedron representing the parameter imprecision increases with the number of criteria. For example, after 25 queries, the number of extreme points is approximately equal to 4500 for problems involving 4 criteria on average. 

We also observe that the timeout is exceeded when applying Two-Phase$_\delta$ on instances with 5 criteria and more. This is due to the fact that the linear program used to solve PMR-optimization problems involves a quadratic number of variables (one per criterion, and one per pair of criteria). 
ILS is the method achieving the best results in terms of error, but it asks too many queries and is very long compared to RIGA. 
For example, with $n=5$ criteria, RIGA asks $13$ queries, ends after $8$ seconds, and its error is below $0.5\%$, whereas ILS needs $39$ queries and $516$s to obtain an error slightly smaller (around $0.1\%$).  

When comparing RIGA to NEMO, we observe that the number of queries and the computation time are quite similar, but NEMO returns a solution of lower quality.

\medskip

\subsection{Multi-Objective Traveling Salesman Problem}

In this section, we consider symmetric Euclidean instances\footnote{\url{https://eden.dei.uc.pt/~paquete/tsp/}} of the MTSP with $50$ cities, and $n=3$ to $6$ objectives. Recall that RIGA generates new solutions using an existing solving method designed for the problem with known weights. For the MTSP, solving the problem with precise weights can be computationally prohibitive. Therefore, we propose instead to use heuristic methods to obtain solutions of good quality more efficiently. For the weighted sum model, we use the Lin-Kernighan heuristic\footnote{\url{http://www.math.uwaterloo.ca/tsp/concorde}} which is one of the best heuristics for solving the single-objective TSP~\cite{Lin1973,ApplegateCR03}. For OWA and Choquet integrals, we propose instead to apply a simple local search based on 2-opt swaps which selects solutions within neighborhoods using the known preference parameters. 
RIGA implemented with these heuristics will be denoted by RIGA$_H$ hereafter. 

\begin{table*}[h!]
\centering{\footnotesize
\scalebox{0.9}{

    \begin{tabular}{  c  c | rrr|rrr| rrr } 
    \cline{1-11} \multicolumn{1}{c}{} & \multicolumn{1}{c|}{}  &  \multicolumn{3}{c|}{\textbf{WS}} &   \multicolumn{3}{c|}{\textbf{OWA}} & \multicolumn{3}{c}{\textbf{Choquet}}   \\  

      \textbf{method} & \bm{$n$} & \textbf{time} & \textbf{queries} & \textbf{error} & \textbf{time} & \textbf{queries} & \textbf{error}  &  \textbf{time} & \textbf{queries} & \textbf{error}  \\ \hline

        \multirow{2}{*}{RIGA}
             & 3 & 66.17  & 14.8  & 0.00 & 546.86 & 5.5  & 0.00 & 235.63  & 23.3 & 0.60      \\
               & 5 & 68.35  & 22.6  & 0.16 & / & /  & / & /  & / & /     \\

          \cline{1-11}
        \multirow{2}{*}{RIGA$_{H}$}
             & 3 & 11.57  & 14.9   & 0.01 & 5.80 & 4.1  & 0.91  & 15.58  & 23.6 & 0.69      \\
             & 5 & 10.45  & 22.2   & 0.10 & 3.75 & 2.7  & 0.46  & 16.67  & 27.9 & 0.80      \\
          \cline{1-11}
            
    \end{tabular}
    }
}
\vspace{-0.2cm}
\caption{\footnotesize Results obtained by RIGA and RIGA$_{H}$ with $\delta=0$, $M=20$, $S=40$ and $K=5$. 
\label{tabComRL-E}}
\end{table*}

Table \ref{tabComRL-E} shows the results obtained by RIGA and RIGA$_H$ on the same instances. 
As expected, RIGA is very slow and the timeout is exceeded when considering 5 criteria or more for OWA and Choquet. We also observe that the computation time is drastically reduced when using heuristics instead of exact methods. 
For example, for instances with 3 criteria, RIGA$_H$ is 6 times faster than RIGA for WS, 109 times faster for OWA and 15 times for Choquet. 
Moreover, the errors obtained by RIGA$_H$ are only slightly higher than that of RIGA, and the number of queries are almost the same. 
Therefore, RIGA$_H$ clearly outperforms RIGA for the MTSP.


\subsubsection{RIGA: parameter setting and evaluation}

In order to determine the best parameter values for RIGA$_H$, we run the algorithm with different the following parameter values: $M =10, 20, 30$, $S = 20, 40, 60$, and $K = 2, 5, 10$. We consider all possible combinations of parameter values, and the average results are given in Table~\ref{tabComTSP}. Note that, since the MTSP is more complicated than the MKP, we had to consider populations of a larger size and perform more iterations to obtain the best results. 

\begin{table*}[h]
\centering{\footnotesize
\scalebox{0.9}{

    \begin{tabular}{  c | c | c | rrr|rrr| rrr } 
    \cline{1-12} \multicolumn{2}{c}{} &  &  \multicolumn{3}{c|}{\textbf{WS}} &   \multicolumn{3}{c|}{\textbf{OWA}} & \multicolumn{3}{c}{\textbf{Choquet}}   \\  

     \multicolumn{1}{c}{\bm{$M$}} & \multicolumn{1}{c}{\bm{$S$}} & \bm{$K$} & \textbf{time} & \textbf{queries} & \textbf{error} & \textbf{time} & \textbf{queries} & \textbf{error}  &  \textbf{time} & \textbf{queries} & \textbf{error}  \\ \hline

        \multirow{9}{*}{$10$}
            & \multirow{3}{*}{$20$} & 2 & 4.85    & 12.9 & 1.34  &  0.98  & 2.2 & 0.49  &  5.26 & 18.0   & 2.59      \\ 
            & & 5  &  5.31   & 12.8 & 1.40  & 1.32   & 2.3    & 0.50 &   5.12     & 18.4    & 3.10            \\ 
            & & 10 &  5.76   & 13.1  & 1.53   &  1.13  & 2.4    & 0.51 &  4.89      & 17.6    & 3.00            \\ \cline{2-12}

            & \multirow{3}{*}{$40$} & 2 &   5.48  & 13.0 & 1.81   & 1.95   & 2.2   & 0.46  &  10.74 & 18.2   & 1.93       \\ 
            & & 5  &  5.88   & 12.9   & 3.52  &  2.00  & 2.7    & 0.47 &   10.47     & 18.0    & 1.70            \\ 
            & & 10 &  6.32   & 13.1   & 3.22  &  2.12  & 2.6    & 0.48  &   9.97     & 17.8    &  1.87           \\ \cline{2-12}
      
            & \multirow{3}{*}{$60$} & 2 &  7.57   & 12.8 & 1.10 &  2.38  & 2.4   & 0.46  & 14.45  & 17.6   &    1.44   \\ 
            & & 5  & 7.94  & 12.9   & 1.20  &  2.80    & 3.1 & 0.46    &   15.44 &  18.5   & 1.72         \\ 
            & & 10 & 9.12  & 12.8   & 1.54  & 2.77   & 2.9    & 0.46 &   13.81     & 17.1 & 1.63            \\  \hline

        \multirow{9}{*}{$20$}
            & \multirow{3}{*}{$20$} & 2 &  5.31   & 22.3 & 0.22  & 2.05  & 2.3 & 0.47  & 9.56  & 27.8   &1.10       \\ 
            & & 5  &   5.81  &  22.2  & 0.29  &  2.00  & 2.7    & 0.48 &  10.20      & 28.5    & 1.17            \\ 
            & & 10 &  7.12   &  22.3  & 0.43  &  2.28  &  2.8  & 0.48 &   8.94     & 27.3    & 1.28            \\ \cline{2-12}

            & \multirow{3}{*}{$40$} & 2 &  9.68   & 22.4 & 0.13  &  3.23  & 2.5 &  0.46 & 15.48 & 23.7   & 0.67       \\ 
            & & 5  &   10.45  &  22.2  & 0.10  & 3.75 & 2.7  & 0.46 &  16.64      & 27.9    & 0.80           \\ 
            & & 10 &  12.96   &  22.8  & 0.22  &  3.82  & 3.4    & 0.47 &  17.39      & 28.4    &      1.03       \\ \cline{2-12}
      
            & \multirow{3}{*}{$60$} & 2 &  14.84   & 22.6 & 0.14  &  5.01  & 2.8 & 0.46  &  25.37 & 28.1   & 0.92       \\ 
            & & 5  & 15.34 & 22.5 & 0.14  &  4.95   & 3.9 & 0.46    &   25.62 & 27.5    & 0.82            \\ 
            & & 10 & 15.79    &  22.7  & 0.07  &  5.45  & 4.1  & 0.46  &  26.01      & 27.8    &1.01         \\  \hline

        \multirow{9}{*}{$30$}
            & \multirow{3}{*}{$20$} & 2 &   7.34  & 28.2 & 0.04  &  2.68  & 2.6 & 0.46  & 12.15  & 38.4   & 0.83       \\ 
            & & 5  &   7.98  &  28.1  & 0.06  & 3.02   & 2.9    & 0.47 &  12.08      & 38.4    & 0.82            \\ 
            & & 10 &  9.60   & 29.7   & 0.21  & 3.50   & 2.7    & 0.46 &  12.40      & 37.7    & 0.92            \\ \cline{2-12}

            & \multirow{3}{*}{$40$} & 2 &  13.48   & 28.8 & 0.15  &  4.92  & 3.2   & 0.46  & 23.13  &    37.7 & 0.58      \\ 
            & & 5  &  14.98   & 29.0   & 0.06  & 5.26  & 3.9 & 0.46 &   23.99     & 37.9     & 0.66             \\ 
            & & 10 & 17.50    & 30.3   & 0.06   &  6.03  & 3.9    & 0.46  &   25.75     & 37.5    & 0.66            \\ \cline{2-12}
      
            & \multirow{3}{*}{$60$} & 2 &  21.23   & 29.4 &  0.02 & 8.62  & 2.5  & 0.46   &  36.05 &    38.3 & 0.51       \\ 
            & & 5  & 22.32 & 29.5 & 0.08  &   7.79   & 4.1 & 0.46  &  36.19     & 37.5   & 0.61           \\ 
            & & 10 & 24.02 & 30.2 & 0.08  &  8.68  & 4.9 & 0.46 &     37.04  & 38.1   & 0.65         \\  \hline

    \end{tabular}
    }
}
\vspace{-0.2cm}
\caption{\footnotesize Results obtained by RIGA$_H$ with $n=5$, $\delta=0$, and $\mu = 0.5$.
\label{tabComTSP}}
\end{table*}

In Table~\ref{tabComTSP}, we observe that setting $M=20$, $S=40$, and $K=5$ gives the best compromise in terms of computation times, number of queries and error. In particular, we indeed observe that the error becomes too high when decreasing the number of generations (i.e. $M=10$), while both the number of questions and the computation times become prohibitive when increasing the number of generations (i.e. $M=30$). We also observe that RIGA$_H$ is too slow when considering $S=60$ without reducing the error significantly. 

For the mutation rate and the tolerance threshold, we have performed similar experiments as that of the MKP, and we have reached the same conclusion: setting $\mu$ to $0.5$ and $\delta$ to $0.5$ is the best choice.


We have compared RIGA$_H$ to RIGA$_S$ which consists in applying genetic operators on solutions (instead of preference parameters), as explained in Subsection 5.4.
We observe that RIGA$_S$ returns solutions of very low quality, the error being above 20\% even when considering $3$ criteria only. To reduce its error, one needs to increase the number of generations, but then it asks too many queries to reach the same level of quality as RIGA$_H$. 
To give an example, for $n=6$ criteria, we managed to obtain an error below $2\%$ with $M=1000$ generations, $S=300$ and $K=30$, but $75$ queries were generated during the execution. 
Similarly, we have observed that RIGA$_H$ outperforms RIGA$_{KCSS}$ implemented with heuristics (see Subsection 5.4 for the description of RIGA$_{KCSS}$).

\subsubsection{Comparisons to other methods}

We now compare RIGA$_H$ to all the other methods considered in the paper, and the results are given in Table \ref{tabCompTSP}.

\begin{table*}[h]
\centering{\footnotesize
\scalebox{0.9}{

    \begin{tabular}{ c  c | rrr|rrr| rrr } 
    \cline{1-11} \multicolumn{1}{c}{} & \multicolumn{1}{c|}{} &  \multicolumn{3}{c|}{\textbf{WS}} &   \multicolumn{3}{c|}{\textbf{OWA}} & \multicolumn{3}{c}{\textbf{Choquet}}   \\  

     \textbf{method} & \bm{$n$} & \textbf{time} & \textbf{queries} & \textbf{error} & \textbf{time} & \textbf{queries} & \textbf{error}  &  \textbf{time} & \textbf{queries} & \textbf{error}  \\ \hline

        \multirow{4}{*}{RIGA$_{H}$}
             & 3 & 11.57  & 14.9   & 0.01 & 5.80 & 4.1  & 0.91  & 15.58  & 23.6 & 0.69      \\
             & 4 & 11.67  & 19.9   & 0.07 & 4.27 & 4.2  & 1.22  & 13.30  & 26.9 & 0.58     \\
             & 5 & 10.45  & 22.2   & 0.10 & 3.75 & 2.7  & 0.46  & 16.67  & 27.9 & 0.80      \\
             & 6 & 11.74  & 24.0   & 0.33 & 5.53 & 5.7 & 0.94   & 20.32 & 32.2 & 1.06  \\  \cline{1-11}
            
        \multirow{4}{*}{NEMO}
             & 3 & 113.23  & 20.0 & 4.20 & 49.81 & 20.0 & 3.81 & 60.13 & 20.0 & 2.53  \\
             & 4 & 144.35  & 20.0 & 5.05 & 39.96 & 20.0 & 4.39 & 51.02 & 20.0 & 3.07  \\
             & 5 & 137.06  & 20.0 & 8.32 & 44.12 & 20.0 & 3.87 &  53.46 & 20.0 & 3.25  \\
             & 6 & 149.40  & 20.0 & 9.50 & 53.75 & 20.0 & 4.09 & 58.29 & 20.0 & 2.89      \\  \cline{1-11}  

        \multirow{4}{*}{IEEP$_{\delta}$}
             & 3 & 10.24  & 7.2   & 0.19 & 15.03 & 5.0 & 0.00 & 152.68 & 11.7 & 1.25     \\
             & 4 & 24.20  & 12.4  & 0.19 & 23.67 & 5.1 & 0.00 & / & / & / \\
             & 5 & 55.86  & 17.8  & 0.22 & / &  / & / & / & / & /     \\
             & 6 & 173.47 & 23.4  & 0.21  & / & / & / & / & / & /     \\  \cline{1-11}
                    
        \multirow{4}{*}{Two-Phase$_{\delta}$}
             & 3 & 140.45  & 12.6 & 0.42 & 59.94 & 4.7 & 0.39  & 375.38 & 17.6 & 0.73     \\
             & 4 & 190.61  & 15.6 & 1.28 & 40.19 & 2.4 & 2.63  & 1074.68 & 53.4 & 1.70     \\
             & 5 & 344.10  & 24.8 & 2.13 & 55.90 & 3.0 & 2.51  & / & / & /     \\
             & 6 & 485.36  & 31.1 & 2.37 & 81.51 & 3.2 & 4.74  & / & / & /     \\  \cline{1-11}
                    
        \multirow{4}{*}{ILS}
             & 3 & 4.14  &  13.7  & 2.12 & 2.98 & 3.5 & 0.99 & 7.61   & 25.7 & 2.98      \\
             & 4 & 10.70 &  19.8  & 2.01 & 3.01 & 4.8 & 1.17 & 35.67  & 51.4 & 4.06     \\
             & 5 & 26.07 &  28.7  & 1.68 & 3.84 & 4.2 & 0.67 & 184.39 & 94.4  & 4.73     \\
             & 6 & 28.57 &  34.1  & 1.69 & 4.32 & 3.7 & 1.36 & 772.55 & 152.8 & 4.81     \\  \cline{1-11}
        
    \end{tabular}
    }
}
\vspace{-0.2cm}
\caption{\footnotesize Results obtained by RIGA$_H$, NEMO, IEEP$_\delta$, Two-Phase$_\delta$ and ILS. 
\label{tabCompTSP}}
\end{table*}

\paragraph{Results obtained with the weighted sum model} For IEEP$_\delta$ and Two-Phase$_\delta$, we set $\delta=0.01$ to allow for the same error as RIGA$_H$. 

First, we observe that RIGA$_H$ clearly outperforms NEMO, Two-Phase$_{\delta}$ and ILS, as it is better than them on all criteria (time, queries, and error). 

Then, when comparing RIGA$_{H}$ to IEEP$_{\delta}$, we see that IEEP$_{\delta}$ asks less queries but its error is higher on the smallest instances.  
On the biggest instance, the number of queries and the error are almost the same, but RIGA$_{H}$ is significantly faster (15 times faster for $n=6$ criteria). 

\paragraph{Results obtained with the OWA model} 
Here we set $\delta=0.02$ for IEEP$_\delta$ and Two-Phase$_\delta$ for the same reasons as before. 

Here also we see that all the methods but IEEP$_{\delta}$ are clearly outperformed by RIGA$_H$. IEEP$_{\delta}$ performs quite well on the smallest instances ($n=3,4$) in terms of error and number of queries, but it is much slower than RIGA$_H$ (e.g., almost $6$ times slower for $n=4$). Moreover, on the largest instances, we observe that IEEP$_{\delta}$ is limited by the use of an exact solving method, which is necessary for its guarantee of performance (the timeout is already exceeded with $n=5$ criteria). 

\paragraph{Results obtained with the Choquet integral model}  For IEEP$_\delta$ and Two-Phase$_\delta$, we set $\delta$ to $0.02$ for the same reasons as before. In the table, 
RIGA$_H$ is clearly better than all other methods in terms of computation time, number of queries, and error. Not surprisingly, we observe here also that IEEP$_\delta$ and Two-Phase$_\delta$ exceed the timeout when increasing the number of criteria (see Section 5.4.1). 

Thus we can conclude that our algorithm is the best one for solving the MTSP overall. In fact, it can be used to solve bigger instances, as shown in Table \ref{tab300TSP} with 300 cities.



\medskip
\begin{table*}[h!]
\centering{\footnotesize
\scalebox{0.9}{

    \begin{tabular}{ c | c | rrr|rrr| rrr } 
    \cline{1-11} \multicolumn{1}{c|}{} & \multicolumn{1}{c|}{}&  \multicolumn{3}{c|}{\textbf{WS}} &   \multicolumn{3}{c|}{\textbf{OWA}} & \multicolumn{3}{c}{\textbf{Choquet}}   \\  

      \multicolumn{1}{c|}{}  & \bm{$n$} & \textbf{time} & \textbf{queries} & \textbf{error} & \textbf{time} & \textbf{queries} & \textbf{error}  &  \textbf{time} & \textbf{queries} & \textbf{error}  \\ \hline

        \multirow{4}{*}{RIGA$_{H}$}
             & 3 &   130.80  & 17.92   &  0.04  & 7.56 & 2.6  & 1.33  &  11.48 & 23.9 & 0.59       \\
             & 4 &   171.14  & 20.10   &  0.22  & 5.57  & 4.1  & 1.11  & 15.56  & 26.9 & 0.80       \\
             & 5 &   178.23  & 21.88   &  0.62  & 2.36 & 3.0  & 0.72  &  18.72 & 27.4 & 1.02       \\
             & 6 &   185.27  & 24.06   &  1.12  & 6.19 & 2.6  & 1.37  & 22.78  & 32.4 & 1.34       \\  \cline{1-11}
        
    \end{tabular}
    }
}
\vspace{-0.2cm}
\caption{\footnotesize Results obtained by RIGA$_H$ with 300 cities.
\label{tab300TSP}}
\end{table*}

\section{Conclusion}

We have proposed an interactive genetic algorithm combined with regret-based incremental elicitation to solve multi-objective combinatorial optimization problems. This new method (called RIGA for Regret-Based Interactive Genetic Algorithm) has been applied to two multi-objective combinatorial problems: knapsack and traveling salesman problems. According to three performance indicators (running time, number of queries and error), RIGA outperforms several existing methods, for different aggregation functions (linear and non-linear). Moreover, RIGA runs in polynomial time and asks no more than a polynomial number of queries. Despite the good experimental results obtained by RIGA (error always less than 1.25\%), RIGA does not have any theoretical performance guarantee. We plan thus to integrate some performance guarantees into RIGA, for example by combining RIGA and IEEP or by studying some particular theoretical properties of the different aggregation functions used.  

\bibliographystyle{plain}
\bibliography{ejor.bib}

\end{document}